\pdfoutput=1

\documentclass[11pt]{article}

\usepackage{acl}

\usepackage{times}
\usepackage{latexsym}

\usepackage[T1]{fontenc}

\usepackage[utf8]{inputenc}

\usepackage{microtype}

\usepackage{inconsolata}

\usepackage{graphicx}

\usepackage{amsmath}
\usepackage{amsthm}
\usepackage{amsfonts}
\usepackage{enumerate}
\usepackage{xspace}
\usepackage{ifthen}
\usepackage{tabularx}
\usepackage{amssymb}
\usepackage{soul}

\newcommand\citepnc[3][]{(#1\citeauthor{#2}, \citeyear{#2}#3)}

\newcommand{\todop}[1]{{\color[HTML]{F5252C} \textbf{(TODO: #1)}}}
\newcommand{\todocite}[1]{\todop{CITE\ifthenelse{\equal{#1}{}}{}{ #1}}}

\newcommand{\gcheck}[0]{{\color[HTML]{196315} \hspace{0mm}$\checkmark$}}
\newcommand{\rex}[0]{{\color[HTML]{A30B16} \hspace{0mm}$\text{\sffamily X}$}}

\title{Procedural Environment Generation for Tool-Use Agents}

\author{Michael Sullivan \And Mareike Hartmann \\
  Department of Language Science and Technology \\
  Saarland Informatics Campus \\
  Saarland University, Saarbrücken, Germany \\
  \texttt{\{msullivan, mareikeh, koller\}@coli.uni-saarland.de} 
  \And Alexander Koller
}

\begin{document}
\maketitle
\begin{abstract}
Although the power of LLM tool-use agents has ignited a flurry of recent research in this area, the curation of tool-use training data remains an open problem\textemdash especially for online RL training. Existing approaches to synthetic tool-use data generation tend to be non-interactive, and/or non-compositional. We introduce RandomWorld, a pipeline for the procedural generation of interactive tools and compositional tool-use data. We show that models tuned via SFT and RL on synthetic RandomWorld data improve on a range of tool-use benchmarks, and set the new SoTA for two metrics on the NESTFUL dataset. Further experiments show that downstream performance scales with the amount of RandomWorld-generated training data, opening up the possibility of further improvement through the use of entirely synthetic data.
\end{abstract}

\section{Introduction}
\label{sec_intro}

A substantial amount of current research has focused on equipping LLMs with the means to employ tools\textemdash external functions (e.g.\hspace{1mm}APIs) that provide the LLM with enhanced knowledge and/or capabilities\textemdash with the goal of achieving AI assistants capable of executing complex sequences of tool calls to accomplish tasks such as information retrieval \citep[e.g.][]{zheng-etal-2024-openresearcher,li2025search,zheng2025deepresearcher}, making online purchases \citep[e.g.][]{yao2022webshop,cai2025large}, editing a user's local files \citep[e.g.][]{trivedi-etal-2024-appworld}, etc. In line with broader findings that LLMs post-trained via reinforcement learining (RL) exhibit superior generalization capabilities to supervised-fine-tuned (SFT) agents \citep{chu2025sft}, recent work demonstrates that LLMs fine-tuned for tool use through online RL can better adapt to tools and tasks not seen during training \citep{qian2025toolrl,feng2025retool}. 

\begin{figure}[t]
\centering
\includegraphics[width=77mm, height=83mm]{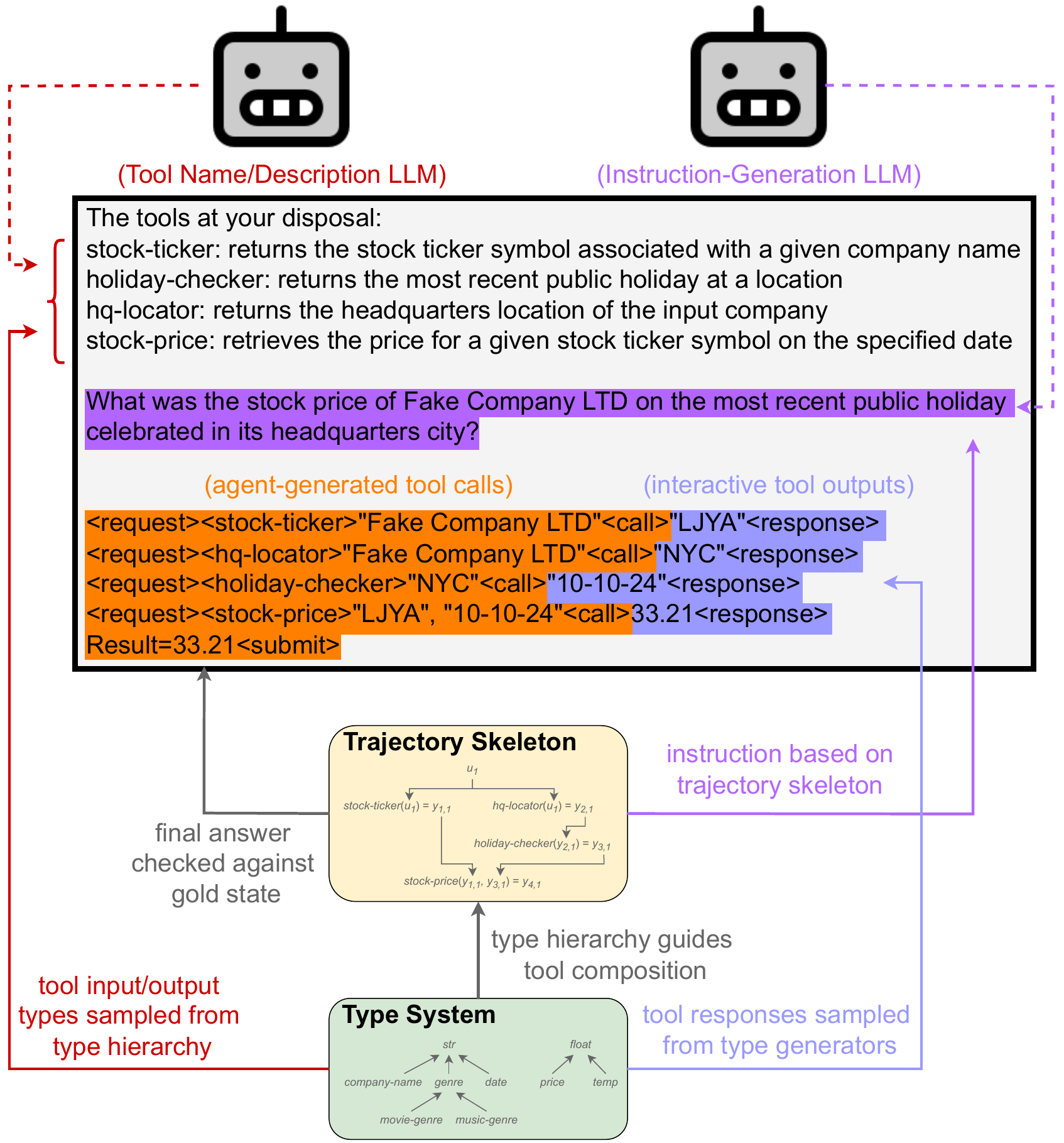}
\caption{Example of an agent executing a (simple) non-linear, compositional task with interactive tools in a RandomWorld environment, with an illustration of the pipeline components that go into the generation of the task and environment.}
\label{fig_fig1}
\end{figure}

To enable the agent to effectively employ tools across a wide range of tasks and domains, LLM tool-use training necessitates a sufficiently rich environment. However, dataset curation for online RL tool-use training remains a major challenge: it is not feasible to directly train tool use agents in the real-world (due to the obvious dangers of such an approach), and hand-crafting simulated APIs\textemdash and tasks using these APIs\textemdash is a time-consuming process that is not feasible at the scale required for (SFT or RL) LLM fine-tuning. As a result, many existing datasets are comprised of either small or non-interactive (i.e.\hspace{1mm}not callable) tool inventories, and/or non-compositional tasks (see Section \ref{sec_related_work}). While training on such datasets may be sufficient to conquer simpler benchmarks \citep[e.g.][etc.; see Section \ref{sec_related_work}]{berkeley-function-calling-leaderboard,patil2024gorilla}, they fail to satisfactorily improve model performance on benchmarks with more complex tasks: namely, tasks necessitating the non-linear chaining of tool calls in an interactive setting \citep[e.g.][etc.; see Sections \ref{sec_related_work} and \ref{sec_experiments}]{zhuang2023toolqa,basu2024nestful,trivedi-etal-2024-appworld}.

On the other hand, those datasets that do contain large, interactive tool inventories with compositional tasks are carefully hand-crafted \citep[e.g.][]{trivedi-etal-2024-appworld}, and so contain a very limited number of tasks: while valuable as benchmarks, these datasets are not suitable for large-scale RL training.

In this paper, we introduce \textit{RandomWorld} (Section \ref{sec_randomworld}), a pipeline for the procedural generation of tools and tasks, that is capable of generating a virtually unlimited amount of training data for online RL (or SFT) training of tool-use agents. As illustrated in Figure \ref{fig_fig1}, RandomWorld employs a fine-grained type hierarchy to generate non-linear DAGs (trajectory skeletons) of tool calls from the signatures of fully-interactive (i.e.\hspace{1mm}callable) synthetic tools (see Section \ref{sec_randomworld}). These trajectory skeletons parameterize the construction of environments: data structures consisting of tools, a goal state (value to return), and an instruction. This procedural generation pipeline results in environments that have:

\begin{enumerate}[i.]
    \item \textit{Depth} (of the tool inventory): a large toolset across a diverse assortment of domains, to facilitate the agent's ability to generalize to unseen tools.
    \item (Non-linear) \textit{Compositionality}: chainable tools\textemdash and objectives necessitating non-linear tool-chaining\textemdash to emulate complex, real-world tasks (see e.g.\hspace{1mm}Figure \ref{fig_fig1}).
    \item \textit{Interactivity}: intermediate tool outputs that are visible to the agent, allowing the inspection of outputs and (if necessary) correction of the tool-call sequence\textemdash as is possible in many real-world settings.
\end{enumerate}

Models fine-tuned with RandomWorld\textemdash through either RL or SFT\textemdash exhibit increased performance on various tool-use benchmarks (Section \ref{sec_experiments}), and set the new SoTA on two NESTFUL \cite{basu2024nestful} metrics. Further experiments show that downstream performance scales with the size of the tool inventory and number of tasks in the training set (Section \ref{sec_scalability}), indicating that further training with RandomWorld can further improve performance, without the need for costly human annotation.

We release all code for the RandomWorld pipeline and the experiments conducted in this paper on GitHub\footnote{\href{https://github.com/coli-saar/randomworld}{https://github.com/coli-saar/randomworld}}.


\section{Related Work}
\label{sec_related_work}

\paragraph{Non-Interactive Tool-Use Datasets.} 

Although many existing tool use datasets have large API inventories and complex tasks that require non-linear tool chaining, they do not include interactive tools (see Table \ref{table_ds_comp}). For example, while APIGen \cite{liu2024apigen} employs real-world, callable APIs during dataset generation, this pipeline only creates non-interactive SFT \citep[and similar, e.g. DPO;][]{rafailov2023direct} training data: namely, queries/instructions annotated with solution paths. Similarly, the ToolBench \cite{qin2024toolllm} dataset contains an interactive test set, but a non-interactive train set\textemdash while it may be theoretically possible to adapt ToolBench to online RL training, the latency accompanying this dataset's use of real-world APIs would impede such an approach. 

In the ToolACE \cite{liu2024toolace} and APIBank \cite{li-etal-2023-api} training sets, both the tasks/solution paths \textit{and} tools are synthesized, as in RandomWorld. Unlike RandomWorld, the tools in ToolACE and APIBank are not callable: an LLM simulates outputs during dataset creation. This again results in SFT-only training sets.

The UltraTool dataset \cite{huang-etal-2024-planning-creation} contains LLM-synthesized tasks and solution paths that require the complex, non-linear compositional use of over 2,000 tools. However, these tools are not functional, and so are not interactive. 

While APIBench \cite{patil2024gorilla} has a deep tool inventory\textemdash over 700 real-world APIs\textemdash it is not interactive or compositional: the LLM-synthesized tasks in this dataset require only one API call.

The non-interactivity of the tools in these datasets impairs evaluation, which further hinders their employment for online RL training, as evaluation and reward are inextricably linked. When tools are not interactive, scores must be assigned through either exact match of the tool call sequence\textemdash which ignores the possibility of a task having more than one solution path\textemdash or LLM-as-a-judge evaluation.

\paragraph{Interactive Tool-Use Datasets and Benchmarks.}

\begin{table*}[t]
\centering
\begin{tabular}{l||ccc|cc}
 & \multicolumn{3}{c|}{\textbf{Tool Inventory}} & \multicolumn{2}{c}{\textbf{Tasks}} \\
\textbf{Dataset} & \textbf{Interactive} & \textbf{Deep} & \textbf{Synthetic} & \textbf{Compositional} & \textbf{Synthetic} \\
\hline
\hline
APIGen & \rex & \gcheck & \rex & \gcheck & \gcheck \\
ToolBench & \rex & \gcheck & \rex & \gcheck & \gcheck \\
UltraTool & \rex & \gcheck & \rex & \gcheck & \gcheck \\
ToolAce & \rex & \gcheck & \gcheck & \gcheck & \gcheck \\
APIBank & \rex & \gcheck & \gcheck & \gcheck & \gcheck \\
APIBench & \rex & \rex & \rex & \rex & \gcheck \\
WebShop & \gcheck & \rex & \rex & \gcheck & \rex \\
BFCL V3 & \gcheck & \rex & \rex & \hspace{1.2mm}\gcheck$^*$ & \hspace{2.5mm}\gcheck$^+$ \\
AppWorld & \gcheck & \gcheck & \rex & \gcheck & \rex \\
\textbf{RandomWorld} & \gcheck & \gcheck & \gcheck & \gcheck & \gcheck 
\end{tabular}
\caption{Comparison between RandomWorld and existing tool-use datasets and benchmarks. ($^*$compositional but linear-only; $^+$checked by human annotators)}
\label{table_ds_comp}
\end{table*}

On the other hand, the WebShop \cite{yao2022webshop} dataset is compositional and fully interactive. However, the WebShop tool inventory is shallow (eight actions/tools), which limits the knowledge obtained from training on the dataset to this narrow domain. 

Similarly, the Berkeley Function-Calling Leaderboard \citep[BFCL V3;][]{berkeley-function-calling-leaderboard} contains fully interactive APIs. As in RandomWorld, BFCL V3 task synthesis begins with the tool call sequence, rather than instruction. Although the BFCL V3 generation procedure results in compositional tasks, it does not result in \textit{non-linear} tasks.

AppWorld \cite{trivedi-etal-2024-appworld} is a simulated world of realistic, carefully hand-crafted users, apps, and APIs. This benchmark contains a deep tool inventory of over 450 fully interactive APIs, which are employed in highly complex and non-linearly compositional tasks. But, as they are hand-crafted, the number of tasks in AppWorld is severely limited (750): this benchmark is therefore not suitable for online RL (or SFT) training. 

\paragraph{Other Synthetic Training Data Pipelines.}

However, synthetic data-generation pipelines show promise in other areas. For example, the AGENTGEN pipeline \citet{hu2024agentgen} demonstrates that LLMs trained on synthetic data can markedly improve on PDDL \cite{McDermott1998PDDLthePD} planning tasks\textemdash and even achieve SoTA results on multiple benchmarks. Similarly, \citet{davidson2025orchestrating} find that LLMs trained on synthetic data generated from the SuperGLUE dataset \cite{wang2019superglue} improve in performance over baseline models. These authors additionally show that a sufficiently advanced synthetic-data pipeline is capable of constructing examples that are more complex than their real-world counterparts. 

\citet{matthews2024kinetix} show that the advantages of synthetic data are not limited to SFT: pretraining an RL agent in randomly-generated physics environments can improve performance on actual downstream tasks. In some cases, this pretrained agent can outperform task-specific models. 

\section{RandomWorld}
\label{sec_randomworld}


As discussed in Section \ref{sec_intro}, RandomWorld leverages a fine-grained type system (see Section \ref{sec_randomworld_sub_types}) to achieve the procedural generation of interactive tools, environments, and non-linear, compositional tasks: instruction/goal state pairs that necessitate non-linear sequences of API calls. Potential input/output types are sampled and passed to an LLM to create tool names and descriptions (see Section \ref{sec_randomworld_sub_tool}). The signatures of these generated tools (and, optionally, hand-crafted tools) are used to generate trajectory skeletons, which parameterize environment and instruction generation (see Section \ref{sec_randomworld_sub_env}).

These generated tools and environments are then easily interfaced with existing RL and SFT pipelines (see Section \ref{sec_randomworld_sub_interface}).

\subsection{Type System}
\label{sec_randomworld_sub_types}

To create the type hierarchy $T$, we constructed 73 base types: fine-grained subtypes of strings (e.g.\hspace{1mm}\textit{month-name}, \textit{movie-title}, \textit{address}), integers (e.g.\hspace{1mm}\textit{age}, \textit{year}, \textit{spotify-id}), and floats (e.g.\hspace{1mm}\textit{hotel-rating}, \textit{temperature}, \textit{price}). To faciltate the task generation procedure (see Section \ref{sec_randomworld_sub_env}), we impose further subtype constraints within this set of custom types: for example, \textit{actor-name} is a subtype of \textit{person-name} (which in turn is a subtype of \textit{string}). Our full type hierarchy is given in Figure \ref{fig_type_hierarchy} in the Appendix.

For each of these base types, we craft a description (used for automated tool generation in Section \ref{sec_randomworld_sub_tool}), a \textit{generator}, and a \textit{recognizer}. Type generators create new instances of that type, and are used to produce automatically-generated tool outputs (see Section \ref{sec_randomworld_sub_tool}): for example, the generator for \textit{month-name} simply samples one of the twelve month names, while the generator for \textit{price} samples a float between 1 and 5000, rounded to two decimal points. Type recognizers are boolean-valued functions that check if an object belongs to the type in question, and are used for type-checking agent inputs when it is interacting with the environment (i.e.\hspace{1mm}using the tools). Generators and recognizers for super-types are inherited from their subtypes.

We implemented three type constructors, which allow for the generation of a theoretically unlimited number of types: $\textit{list}\colon T\to T$, which takes a type $t$ and returns the type $\textit{list}(t)$ of lists of objects of type $t$; $\textit{dict}\colon T\times T\to T$, which takes types $t$, $u$ and returns the type $\textit{dict}(t,u)$ of dictionaries mapping objects of type $t$ to objects of type $u$; and $\textit{union}\colon T\times T\to T$, which takes types $t$, $u$ and returns the type $\textit{union}(t,u)$ of objects of type $t$ or $u$. Subtype relations between\textemdash and type recognizers/generators for\textemdash constructed types are automatically inferred from their constituent type(s).

A detailed description of the RandomWorld type system is located in Appendix \ref{app_types}.

\subsection{Tool Creation}
\label{sec_randomworld_sub_tool}

To automatically generate a tool\textemdash i.e.\hspace{1mm}an LLM-callable function\textemdash in RandomWorld, we sample input types $X_1,\dots,X_n$ and output types $Y_1,\dots,Y_m$ (see Section \ref{sec_randomworld_sub_types}). We then use an LLM to generate a possible name and description for a tool $f\colon X_1\times\dots\times X_n\to Y_1\times\dots\times Y_m$, and prompt the LLM to score the plausibility and realism of the generated description on a scale from 1-5: tools with a score less than 4 are discarded.

When passed inputs $x_1,\dots,x_n$, the tool returns $y_1,\dots,y_m$, where each $y_i$ is sampled from the type generator for $Y_i$. While the agent is interacting with the environment, input/output pairs $((x_1,\dots,x_n),(y_1,\dots,y_m))$ are temporarily stored, ensuring that\textemdash from the perspective of the agent\textemdash the tool always returns the same output for a given input (see Section \ref{sec_randomworld_sub_env_sub_env}). If the tool is passed an input of an invalid type\textemdash as determined by tool input type's recognizer (see Section \ref{sec_randomworld_sub_types})\textemdash the tool returns an error message.

In comparison to a tool generation procedure in which an LLM directly codes the tools, our type-guided procedure guarantees the behavior of the tools: each tool always returns values of its annotated type. This consistency facilitates the environment and task generation process of Section \ref{sec_randomworld_sub_env}. Examples of tools generated by RandomWorld are given in Table \ref{table_tool_exs} in the Appendix.

Note that a tool in RandomWorld is simply a function annotated with input/output types (see Section \ref{sec_randomworld_sub_types}). This permits the use of hand-crafted tools in place of\textemdash or alongside\textemdash synthetic tools. For example, our experiments (Section \ref{sec_experiments}) included six hand-crafted calculator tools: \textit{add}, \textit{subtract}, \textit{multiply}, \textit{divide}, \textit{max}, and \textit{min}.

RandomWorld additionally implements dependently-typed tools: tools whose output type is a function of their input. For example, $\textit{add}\colon (t\leq\textit{float})\times(t\leq\textit{float})\to t$ will (for example) take the prices of two items and return their total price (see Figure \ref{fig_trj_ex}). This permits richer and more controlled environment generation (see Section \ref{sec_randomworld_sub_env}). However, due to the added complexity that they introduce, we only experiment with hand-crafted dependently-typed tools.

\subsection{Environment and Instruction Generation}
\label{sec_randomworld_sub_env}

In comparison to most synthetic task generation pipelines, in which an LLM first synthesizes a query that is then used to generate a sequence of API calls \citep[e.g.][]{liu2024apigen,qin2024toolllm,liu2024toolace}, RandomWorld tasks are synthesized by first generating a sequence of API calls through a type-guided sampling procedure (see Section \ref{sec_randomworld_sub_env_sub_trjs}). These API calls are then used to populate the environment (i.e.\hspace{1mm}generate tool input/output values; see Section \ref{sec_randomworld_sub_env_sub_env}) and generate a corresponding instruction via LLM (see Section \ref{sec_randomworld_sub_env_sub_instructions}). 

See Appendix \ref{app_trjs} for examples of RandomWorld-generated instructions and tool call sequences.

\subsubsection{Trajectory Skeletons}
\label{sec_randomworld_sub_env_sub_trjs}

\begin{figure*}[t]
\centering
\includegraphics[width=\textwidth]{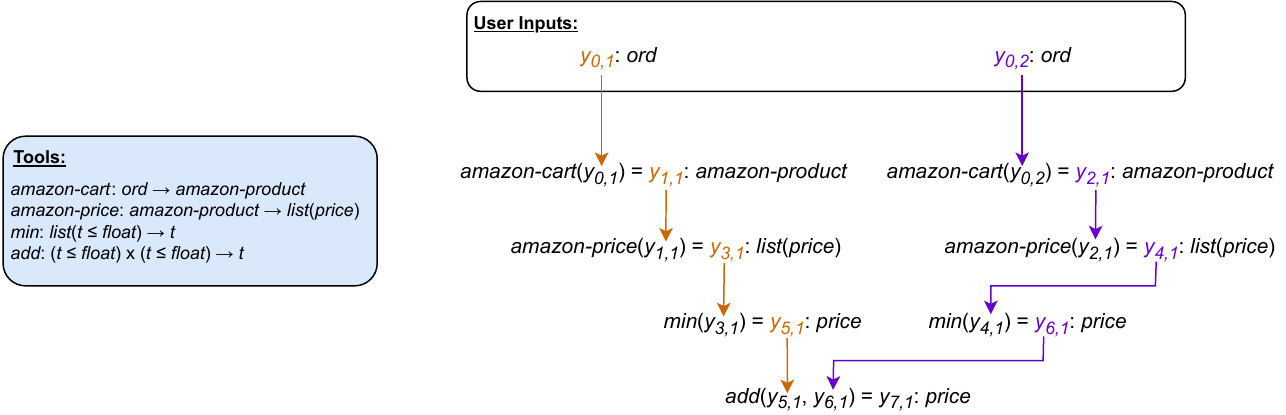}
\caption{Example of a non-linear trajectory skeleton that corresponds to the instruction \textit{``how much will the $y_{0,1}$-th and $y_{0,2}$-th most recently-added items in my Amazon cart cost together, if purchased at the lowest-available price?''}.}
\label{fig_trj_ex}
\end{figure*}

To automatically construct a task\textemdash i.e.\hspace{1mm}an instruction/goal state pair\textemdash in RandomWorld, we first generate a \textit{trajectory skeleton}: a sequence of tool calls $f_1,\dots,f_n$, along with annotations indicating the output(s) of the tool(s) $f_j,\dots,f_k$ that $f_i$ ($i>j,k$) takes as input (see Figure \ref{fig_trj_ex}). The fine-grained type system defined in Section \ref{sec_randomworld_sub_types} permits lazy evaluation: we do not compute tool outputs until the trajectory skeleton is constructed. This is essential to RandomWorld's type-checked function sampling procedure, which necessitates back-and-forth construction and trimming of trajectory skeletons until convergence.

A trajectory skeleton is constructed by first sampling \textit{user input} type(s) $Y_{0,1},\dots,Y_{0,m}$, corresponding to the value(s) that will be fed to the agent in the instruction, rather than as a tool output: e.g.\hspace{1mm}\textit{``find \underline{comedy} movies on Netflix that last less than \underline{two hours}''}. We then sample a trajectory length $\ell$ such that $\ell_\textit{min}\leq\ell\leq\ell_\textit{max}$, where $\ell_\textit{min}$ and $\ell_\textit{max}$ are pre-defined minimum and maximum trajectory lengths, to (for example) control the complexity of training examples for curriculum learning.

Next, we sample a tool $f_1$ that is compatible with the variable set $V$ (the user-input types $Y_{0,1},\dots,Y_{0,m_0}$): i.e.\hspace{1mm}such that for each input type $X_{1,i}$ of $f_1$, there is a type $Y\in V$ with $Y\leq X_{1,i}$. For example, in Figure \ref{fig_trj_ex}, $\textit{amazon-cart}\colon\textit{ord}\to\textit{amazon-product}$ is compatible with the variable set $V=\{y_{0,1}\colon\textit{ord},y_{0,2}\colon\textit{ord}\}$ (as $\textit{ord}\leq\textit{ord}$).

We then record the types in $V$ that match each argument position of $f_1$, and add the output types $Y_{1,1},\dots,Y_{1,m_1}$ of $f_1$ to $V$. In the example in Figure \ref{fig_trj_ex}, we record that the argument of \textit{amazon-cart} is $y_{0,1}$, and add its output type to the variable set: $V\gets V\cup\{y_{1,1}\colon\textit{amazon-product}\}$.

This procedure continues until the trajectory skeleton reaches $\ell$ tool calls; the output(s) of the last tool in the sequence are taken as the goal state.

Let $U(f_\ell)=1$, and for all $i<\ell$, let $U(f_i)=1$ if at least one output of $f_i$ is taken as input by some $f_k$ such that $U(f_k)=1$, and $U(f_i)=0$ otherwise. In words: $U(f_i)=1$ if at least one output of $f_i$ is used (directly or indirectly) to compute the goal state. We remove from the trajectory skeleton all tool calls $f_i$ such that $U(f_i)=0$, and sample new tool calls to add to this trimmed trajectory skeleton until it again reaches length $\ell$. This procedure repeats until the trajectory skeleton contains $\ell$ tool calls, and $U(f_i)=1$ for all $1\leq i\leq\ell$. 

When generating a set of trajectory skeletons\textemdash e.g.\hspace{1mm}to construct a training or evaluation set\textemdash we filter out all duplicate trajectory skeletons before environment generation (Section \ref{sec_randomworld_sub_env_sub_env}), in order to ensure diversity.

\subsubsection{Environments}
\label{sec_randomworld_sub_env_sub_env}

Once the trajectory skeleton has been generated, we sample user input values and compute output values for each tool call in the sequence, thereby populating an \textit{environment}: a data structure that contains tool input/output values, user information (see below), a goal state, and an instruction (see Section \ref{sec_randomworld_sub_env_sub_instructions}). The storage of each tool's input/output values as they are computed ensures deterministic outputs from the agent's perspective.

We optionally allow tools to be (manually) assigned to an app: when a tool is assigned to an app, the agent must login to that app before using the tool. A random username and password is generated for each unique app associated with a trajectory skeleton. 

With probability $p_\textit{g}$ (a tunable parameter), the instruction (see Section \ref{sec_randomworld_sub_env_sub_instructions}) contains the username and password to an app used in the trajectory skeleton. Otherwise, only the username is provided, and the agent must use the \textit{password-manager} tool to retrieve the user's password for that app. 

\subsubsection{Instructions}
\label{sec_randomworld_sub_env_sub_instructions}

After the environment has been populated from a trajectory skeleton, we employ an LLM to generate an instruction for that environment. We prompt the LLM with descriptions of each tool in the trajectory, the value(s) of the user input(s), and the trajectory skeleton\textemdash to ensure that the LLM does not leak information about any of the tool outputs into the instruction, we replace all non-user-input values with variables of form ``x\_i,j''.

As the instruction-generation LLM can create instructions that do not provide sufficient information to reach the goal state\footnote{e.g.\textit{``perform a series of arithmetic operations''} when there are many calculator tool calls in the trajectory.}, we verify each generated instruction by checking whether an LLM can reach the goal state using the instruction. As we are merely verifying the informativity of the instructions, we provide the instruction-verification LLM with several advantages not afforded to the agents trained in Section \ref{sec_experiments}: we provide the type signatures of the tools, including descriptions and examples of each input/output type; we provide only the tools required to reach the goal (i.e.\hspace{1mm}we do not include any distractor tools; see Section \ref{sec_randomworld_sub_interface}); we provide tool descriptions in the order that the tool calls are made in the gold trajectory; and we disable the app login mechanism. We discard all environments in which the instruction-verification LLM was unable to reach the goal state given the instruction.

Although we take care to prevent information leakage with respect to tool \textit{outputs}, the instruction generation process outlined here does carry a non-trivial risk of information leakage regarding tool names and/or descriptions, which may inadvertently simplify the tasks (see Appendix \ref{app_info_leakage} for a detailed analysis). However, the results of our RandomWorld-trained models in Section \ref{sec_experiments} indicate that any information leakage that may have occurred did not substantially detract from the models’ performance.

\subsection{Agent Interface}
\label{sec_randomworld_sub_interface}

RandomWorld is fully compatible with TRL \cite{vonwerra2022trl} text environments: for evaluation and RL training, we simply create a text environment for each RandomWorld environment, and pass the agent and the RandomWorld environment's tools to the text environment.

For SFT training, we use the trajectory skeleton and stored tool input/output values to construct a training instance that reaches the goal state. 

Prompts are constructed by prepending a tool inventory to the instruction. The tool inventory presented to the agent consists of all tools used in the corresponding trajectory skeleton, along with a number of randomly-selected distractor tools, as specified by the pre-defined ratio $r_{\textit{dist}}$. 

\section{Experiments}
\label{sec_experiments}

To assess the efficacy of RandomWorld-generated data for SFT and RL, we fine-tuned Llama-3.1-8B-Instruct\footnote{\href{https://huggingface.co/meta-llama/Llama-3.1-8B-Instruct}{https://huggingface.co/meta-llama/Llama-3.1-8B-Instruct}} and Qwen2.5-7B-Instruct \cite{yang2024qwen2} on 12,000 environments generated from a set of six hand-crafted and 550 procedurally-generated tools, with $\ell_\textit{min}=2$, $\ell_\textit{max}=8$, and $r_\textit{dist}=1.0$. 

We employed GPT-4o \cite{hurst2024gpt} as the tool-creation (see Section \ref{sec_randomworld_sub_tool}) and instruction-generation/verification LLM (see Section \ref{sec_randomworld_sub_env_sub_instructions}). We achieved a pass rate rate of $\sim$11\% for tool generation (i.e.\hspace{1mm}$\sim$89\% of the candidate tools were discarded; see Section \ref{sec_randomworld_sub_tool}) and $\sim$60\% for environment generation (including de-duplicated trajectory skeletons; see Section \ref{sec_randomworld_sub_env}). The total OpenAI API cost was roughly \$150 USD.

We then evaluated these RandomWorld-trained models against baseline tool-use models (see Section \ref{sec_experiments_sub_baseline}) on a series of downstream tool-use benchmarks (Section \ref{sec_experiments_sub_eval}).

\subsection{Training}
\label{sec_experiments_sub_training}

For each model, we trained one variant using Group Relative Policy Optimization \citep[GRPO;][]{shao2024deepseekmath} from scratch (i.e.\hspace{1mm}``zero RL''), and one with standard SFT. The GRPO variants were trained for one epoch with a group size of 8, $\beta=0.04$, a batch size of 32, a learn rate of $10^{-6}$, and a temperature of 0.75. We employed an exact match reward function with respect to the goal state for GRPO training, as in \citet{guo2025deepseek}. A full description of our online RL training regimen is located in Appendix \ref{app_grpo}.

The SFT variants were trained with a batch size of 32, a learn rate of $10^{-5}$, and weight decay of $10^{-3}$, using early stopping when performance failed to improve on a withheld validation set (four epochs for Llama, six for Qwen). For all models, we employed LoRA \cite{hu2022lora} adapters with $r=64$, $\alpha=128$, and 0.05 dropout on all $Q$, $K$, $V$, and $O$ attention projection matrices.

\subsection{Baseline Models}
\label{sec_experiments_sub_baseline}

\begin{table*}[h]
\centering
\scalebox{0.79}{\begin{tabular}{l||l|ll|lllll}
 & \textbf{RandomWorld} & \multicolumn{2}{c|}{\textbf{ToolQA}} & \multicolumn{5}{c}{\textbf{NESTFUL}} \\
\textbf{Model} & \textbf{Test} & \textbf{Easy} & \textbf{Hard} & \textbf{F1 Func.} & \textbf{F1 Param.} & \textbf{Part. Acc.} & \textbf{Full. Acc.} & \textbf{Win Rate} \\
\hline
\hline
$\textbf{Llama}^*$ & 0.483 & 0.236 & 0.015 & 0.53 & 0.38 & \underline{0.24} & \underline{0.19} & \underline{0.16} \\
$\text{ToolACE-8B}^*$ & 0.170 & 0.206 & \underline{0.046} & 0.51 & 0.27 & 0.14 & 0.0 & 0.0 \\
Llama-RW-GRPO (Ours) & 0.721 & \underline{0.251} & 0.023 & \underline{0.89} & \underline{0.62} & 0.22 & 0.17 & 0.10 \\
Llama-RW-SFT (Ours) & \textbf{0.855} & 0.086 & 0.019 & 0.61 & 0.36 & 0.11 & 0.07 & 0.04 \\
\hline
$\textbf{Qwen}^*$ & 0.109 & 0.258 & 0.011 & 0.90 & 0.66 & 0.27 & 0.21 & 0.28 \\
$\text{Hammer2.0-7B}^*$ & 0.011 & 0.005 & 0.001 & 0.61 & 0.46 & \textbf{0.31} & \textbf{0.25} & \textbf{0.34} \\
Qwen-RW-GRPO (Ours) & 0.536 & 0.243 & \underline{0.063} & 0.92 & 0.69 & 0.28 & 0.22 & 0.30 \\
Qwen-RW-SFT (Ours) & \underline{0.714} & \underline{0.307} & 0.019 & \textbf{0.96} & \textbf{0.71} & 0.28 & 0.22 & 0.32 \\
\hline
$\text{Mixtral-8x22B}^*$ & \textemdash & \textemdash & \textemdash & 0.64 & 0.48 & 0.29 & 0.21 & 0.29 \\
$\text{GPT-3.5}^+$ & 0.348 & \textbf{0.368} & \textbf{0.082} & \textemdash & \textemdash & \textemdash & \textemdash & \textemdash
\end{tabular}}
\caption{Results of our models, ToolACE-8B, Hammer2.0-7B, and the base Llama-3.1-8B-Instruct/Qwen2.5-7B-Instruct on the RandomWorld, ToolQA, and NESTFUL benchmarks. The best results in each column are indicated in \textbf{bold}, and the best results within each model type (i.e.\hspace{1mm}Llama or Qwen) are \underline{underlined}. ($^*$NESTFUL results reported in \citeauthor{basu2024nestful}, \citeyear{basu2024nestful}; $^+$ToolQA results reported in \citeauthor{zhuang2023toolqa}, \citeyear{zhuang2023toolqa})}
\label{table_results}
\end{table*}

We compared our fine-tuned Qwen models to Hammer2.0-7B \cite{lin2025robust}, a Qwen2.5-7B-Instruct model fine-tuned via SFT on an augmented version of the xlam-function-calling-60k tool-use dataset \cite{zhang2024xlam}, which was generated using the APIGen pipeline \cite{liu2024apigen} discussed in Section \ref{sec_related_work}. We compared our fine-tuned Llama models to ToolACE-8B \cite{liu2024toolace}, a Llama-3.1-8B-Instruct model fine-tuned (SFT) on the ToolACE dataset (see Section \ref{sec_related_work}).

We additionally compared our models to the base Llama-3.1-8B-Instruct and Qwen2.5-7B-Instruct. 

\subsection{Evaluation}
\label{sec_experiments_sub_eval}

We evaluated the models on a RandomWorld test set: 276 environments ($\ell_\textit{min}=2$, $\ell_\textit{max}=8$) generated from 75 tools not used in training. All models were evaluated with a maximum of 15 turns (tool calls or attempted tool calls), and a maximum of 128 new tokens per turn. As our evaluation metric, we report exact match accuracy with the goal state. 

We additionally evaluated the models on the ToolQA \cite{zhuang2023toolqa} and NESTFUL \cite{basu2024nestful} benchmarks.

\paragraph{ToolQA:} 

a question-answering dataset spanning eight domains and requiring the use of twelve fully-interactive tools, ranging from database-searching tools to a Python interpreter. Models are evaluated on exact match between the gold solution and their returned answer. Questions in ToolQA are sorted into Easy (800) and Hard (730) subsets, based on the complexity of the required tool calls and reasoning process. We evaluated the models using the ReAct framework \cite{yao2023react}, the best-performing approach in \citet{zhuang2023toolqa}.

\paragraph{NESTFUL:} a dataset specifically designed to evaluate LLMs on nested (i.e.\hspace{1mm}chained) sequences of API calls. NESTFUL consists of 1,861 tasks in the domains of mathematical reasoning and generic coding, requiring the use of over 4,000 callable APIs. However, unlike ToolQA and the RandomWorld test set, NESTFUL is not fully interactive: the agent returns only a sequence of API calls, from which the final answer is then computed. 

NESTFUL evaluates models across five dimensions: \textit{F1-Function} (F1F), the F1 score between the predicted and gold API sequences; \textit{F1-Parameter} (F1P), between the predicted and gold parameter names; \textit{Partial Sequence Accuracy} (PSA), the percentage of correct API calls (API and parameter names) with respect to the gold sequence; \textit{Full Sequence Accuracy} (FSA), the percentage of tasks in which the model predicted the entire gold sequence of API calls; and \textit{Win Rate} (WR), the percentage of tasks in which the value returned by the predicted sequence of API calls matches the gold answer. Following \citet{basu2024nestful}, we evaluated our models in a 3-shot in-context learning (ICL) setting.

\subsection{Results and Discussion}
\label{sec_experiments_sub_results}

The results of our experiments are given in Table \ref{table_results}.

\paragraph{Qwen-RW-SFT Achieves NESTFUL F1 SoTA.}

On the NESTFUL benchmark, our Qwen-RW-SFT model sets the SoTA for the F1F (0.96) and F1P (0.71) scores, both of which reflect the correctness of the predicted API calls. In fact, aside from Llama-RW-SFT, all of our models outperform the previous SoTA (Mixtral-8x22B-Instruct-v0.1\footnote{\href{https://huggingface.co/mistralai/Mixtral-8x22B-Instruct-v0.1}{https://huggingface.co/mistralai/Mixtral-8x22B-Instruct-v0.1}}).

\paragraph{Synthetic Data Improves Model Performance.}

Our results show that training tool-use agents purely on procedurally-generated synthetic data improves model performance: Qwen-RW-SFT improves over the base Qwen model on all eight benchmarks/metrics, and Qwen-RW-GRPO on all but one (ToolQA-Easy). Qwen-RW-GRPO also narrows the gap by more than 50\% on ToolQA-Hard between open-source models and the closed-source GPT-3.5 \citep[the current SoTA, as reported in][]{zhuang2023toolqa}, a model that is 25 times larger.

Although Llama-SFT-GRPO only improves over the base Llama model on three benchmarks, Llama-RW-GRPO improves on all benchmarks aside from the NESTFUL PSA, FSA, and WR metrics. In particular, Llama-RW-GRPO nearly doubles the performance of the base Llama model on the RandomWorld test set and NESTFUL F1F/F1P metrics, and performance relative to the base model only degrades slightly for PSA, FSA, and WR.  

Overall, the superior results of Llama-RW-GRPO over Llama-RW-SFT (and ToolACE-8B) demonstrate the utility of data-creation pipeline for tool-use that is compatible with online RL tuning, while the SoTA and near-SoTA results of Qwen-RW-SFT show the importance of a pipeline compatible with both kinds of fine-tuning.

\paragraph{Data Quality Matters.} Both Qwen-RW-GRPO and Qwen-RW-SFT outperform Hammer2.0-7B\textemdash another Qwen2.5-7B-Instruct model trained on much more data (see Section \ref{sec_experiments_sub_baseline})\textemdash on all benchmarks aside from the NESTFUL PSA, FSA, and WR metrics. On those metrics, the performance of Qwen-RW-SFT is on par with that of Hammer2.0-7B\textemdash the current 3-shot ICL SoTA on all three metrics. Our Qwen models achieve the second- and third-best NESTFUL 3-shot ICL WR: the previous second-best (Mixtral-8x22B-Instruct-v0.1) has a score of 0.29.

The fact that both Qwen-RW-SFT and Hammer2.0-7B are trained via SFT on synthetic tool-use data raises the question as to why our model outperforms Hammer2.0-7B, especially considering that Hammer2.0-7B is trained on significantly more data (67,500 examples).

Note that both Hammer2.0-7B and ToolACE-8B are trained on datasets in which an LLM generates queries/instructions from a set of tools, from which the sequence of API calls is then derived (see Sections \ref{sec_related_work} and \ref{sec_experiments_sub_baseline}). On the other hand, in RandomWorld, the sequence of tool calls is already determined, and is merely described by the LLM (see Section \ref{sec_randomworld_sub_env}). This procedure likely generates more complex tasks than an LLM alone is capable of envisioning \citepnc[c.f.\hspace{1mm}the findings of ]{davidson2025orchestrating}{ with respect to synthetic reasoning data; see Section \ref{sec_related_work}}, thereby providing our models with a richer training set than those of the baselines. 

In addition, Llama-RW-GRPO outperforms ToolACE-8B\textemdash another Llama-3.1-8B-Instruct model trained via SFT on a similar amount of data (see Section \ref{sec_experiments_sub_baseline})\textemdash on all benchmarks/metrics but ToolQA-Hard: this result in particular further demonstrates the importance of an RL-capable method such as RandomWorld.

\subsection{Analysis: ToolQA-Easy}
\label{sec_experiments_sub_analysis}

On the ToolQA-Easy dataset, we observe a puzzling phenomenon: GRPO training on RandomWorld improves the performance Llama-3.1-8B-Instruct, but decreases that of Qwen2.5-7B-Instruct, while Qwen2.5-7B-Instruct benefits from SFT training and Llama-3.1-8B-Instruct does not. 

Note that the two models that did not benefit from RandomWorld training with respect to ToolQA-Easy (Llama-RW-SFT and Qwen-RW-GRPO) have the best and worst performance of the RandomWorld-trained models on the withheld RandomWorld test set (respectively), while the two models that did benefit from RandomWorld training have similar performance on RandomWorld Test (0.714 and 0.721). These results indicate that Llama-RW-SFT (0.855 on RandomWorld Test) is overfitting the RandomWorld distribution, while Qwen-RW-GRPO (0.536) is \textit{under}fitting. 

\begin{table*}[t]
\centering
\scalebox{0.95}{\begin{tabular}{ll||l|ll|lllll}
 & & \textbf{RW} & \multicolumn{2}{c|}{\textbf{ToolQA}} & \multicolumn{5}{c}{\textbf{NESTFUL}} \\
\textbf{Tools} & \textbf{Tasks} & \textbf{Test} & \textbf{Easy} & \textbf{Hard} & \textbf{F1 Func.} & \textbf{F1 Param.} & \textbf{Part. Acc.} & \textbf{Full. Acc.} & \textbf{Win Rate} \\
\hline
\hline
100\% & 100\% & \textbf{0.536} & 0.243 & 0.063 & \textbf{0.92} & \textbf{0.69} & \textbf{0.28} & \textbf{0.22} & \textbf{0.30} \\
100\% & 25\% & 0.348 & \textbf{0.250} & 0.056 & \textbf{0.92} & 0.68 & 0.27 & 0.21 & 0.29 \\
25\% & 100\% & 0.529 & 0.231 & \textbf{0.073} & \textbf{0.92} & 0.68 & \textbf{0.28} & \textbf{0.22} & \textbf{0.30} \\
25\% & 25\% & 0.297 & 0.246 & 0.059 & \textbf{0.92} & 0.68 & 0.27 & 0.21 & 0.29
\end{tabular}}
\caption{Scalability study results on the RandomWorld, ToolQA, and NESTFUL benchmarks. The top row (100\% tools, 100\% tasks) corresponds to the Qwen-RW-GRPO model of Section \ref{sec_experiments}/Table \ref{table_results}.}
\label{table_scalability}
\end{table*}

This hypothesis is further supported by Llama-RW-GRPO and Qwen-RW-SFT outperforming Llama-RW-SFT and Qwen-RW-GRPO (respectively) on all nearly all NESTFUL metrics. That Qwen-RW-SFT outperforms Llama-RW-GRPO on ToolQA-Easy is likely due to the higher performance of the base Qwen2.5-7B-Instruct model (0.258) over that of the base Llama-3.1-8B-Instruct (0.236). 

This suggests an optimal early-stopping threshold for validation accuracy during RandomWorld training of $\sim$0.718. Rather than increasing validation performance beyond this limit, additional downstream performance gains can be obtained through increasing the difficulty of the RandomWorld environment, by increasing the trajectory skeleton length ($\ell_\textit{max}$) and/or applying the scaffolded curriculum learning method proposed in Section \ref{sec_limitations}.

\section{Scalability Study}
\label{sec_scalability}

Next, we examined the effect of number of synthetic tools and tasks on downstream performance.

\subsection{Experimental Setup}
\label{sec_scalability_sub_setup}

The first dataset, designed to evaluate the effect of task number, was constructed by randomly removing 75\% of the tasks from the original training set used in Section \ref{sec_experiments}, leaving 25\% remaining. 

The second was designed to evaluate the effect of tool inventory size, and was constructed by removing 75\% of the tools from the original training set, then generating 12,000 unique tasks from that restricted tool set. The third was derived from this dataset by randomly removing 75\% of those tasks.

We fine-tuned Qwen2.5-7B-Instruct with GRPO on these three datasets, using the same hyperparameter configuration as in Section \ref{sec_experiments_sub_training}. We then evaluated the three models on the benchmarks in Section \ref{sec_experiments_sub_eval}, with the same experimental settings.

\subsection{Results and Discussion}
\label{sec_scalability_sub_results}

The results in Table \ref{table_scalability} show that the number of tools and tasks the model is trained on affect downstream performance. These results are most pronounced on the RandomWorld test set: Qwen-RW-GRPO performed worse than the base model on ToolQA-Easy\textemdash explaining the observed \textit{increased} performance when trained on fewer tasks\textemdash and NESTFUL is not fully in-distribution with respect to RandomWorld (as discussed in Sections \ref{sec_experiments_sub_eval} and \ref{sec_experiments_sub_results}), so we would not expect a marked effect on those benchmarks. 

Interestingly, training on fewer tools improves ToolQA-Hard performance: the model tuned with 25\% tools and 100\% tasks improves over Qwen-RW-GRPO by 15\%, further narrowing the gap with the SoTA (0.082; see Table \ref{table_results}). We hypothesize that this is due to the shallow ToolQA tool inventory (12 tools; see Section \ref{sec_experiments_sub_eval}), which may not benefit from training on a wider range of tools. 

On the RandomWorld test set, we observe that the number of tasks has a more substantial impact than number of tools: tuning on 25\% of the number of tools and the same number of tasks as Qwen-RW-GRPO only slightly decreases RandomWorld test set performance. Conversely, tuning on 25\% of the tasks with the same number of tools results in a more sizable decrease. However, with fewer tasks, the importance of tool inventory size appears to increase: the model tuned with 25\% of the tasks and 100\% of the tools outperforms that tuned with 25\% of the tasks and tools by a relatively wide margin.

\section{Conclusion}
\label{sec_conclusion}

We introduced RandomWorld (Section \ref{sec_randomworld}), a pipeline for the generation of virtually unlimited synthetic tools and tool-use training data\textemdash including interactive tools and compositional, non-linear tasks\textemdash for both SFT and online RL training. In Section \ref{sec_experiments}, we showed that models can be fine-tuned on RandomWorld training data to improve performance on a variety of downstream benchmarks. Furthermore, a Qwen2.5-7B-Instruct model fine-tuned solely on synthetic RandomWorld data achieves the new SoTA for two metrics on the NESTFUL tool-use benchmark.

The results of our scalability study (Section \ref{sec_scalability}) demonstrate the importance of synthetic data generation for tool-use agents: training on fewer examples leads to decreased downstream performance. The converse of this finding\textemdash namely, that training on \textit{more} examples leads to \textit{increased} downstream performance\textemdash indicates that the RandomWorld pipeline can continue to be used to further advance the SoTA on tool-use benchmarks, without the need for costly, real-world data curation. 

\section{Limitations}
\label{sec_limitations}

\paragraph{Experiments.}

Although we evaluate two model types (Llama-3.1-8B-Instruct and Qwen2.5-7B-Instruct) and two fine-tuning regimens (SFT and GRPO), our experiments are limited to relatively small 7-8 billion parameter models. In future work, we intend to evaluate a wider range models, in order to verify that the advantages conferred by RandomWorld continue to hold for larger models and a variety of architectural configurations. 

Our experiments are also fairly restricted in scale: those conducted in Section \ref{sec_experiments} involve only 12,000 tasks and 556 tools, primarily due to computational resource limitations. While this is on par with the size of training datasets such as ToolACE \cite{liu2024toolace}, the results of Section \ref{sec_scalability} suggest the importance of training with a larger RandomWorld-generated dataset in future work.

Similarly, we only evaluate our models on three benchmarks. Although our results in Section \ref{sec_experiments} definitively show the superiority of our method with respect to those downstream tasks, we intend to evaluate RandomWorld-trained models on a larger set of benchmarks in future work, in order to further demonstrate the effectiveness of this approach.

Finally, we do not experiment with more advanced training regimens, such as curriculum learning. Given the demonstrated effectiveness of curriculum learning for training tool-use agents with RL \citep[e.g.][]{qi2025webrl} and RandomWorld's built-in compatibility with such methods (see Section \ref{sec_randomworld_sub_env}), future work should incorporate the use of curriculum learning into agent training with RandomWorld. 

\paragraph{Method.}

The effectiveness of RandomWorld's tool- and instruction-generation procedures are limited by the effectiveness of the tool- and instruction-generation LLMs (see Section \ref{sec_randomworld_sub_env}). Similarly, recall of the instruction-verification LLM\textemdash which is intended to filter out insufficiently-informative instructions (as discussed in Section \ref{sec_randomworld_sub_env_sub_instructions})\textemdash is dependent on the choice of model: we have no mechanism to account for cases in which the instruction \textit{is} sufficiently informative, but the instruction-verification LLM is simply unable to complete the task. This may have the undesired effect of unnecessarily filtering out the most difficult tasks.  

This limitation could potentially be mitigated through the use of a scaffolded curriculum learning approach, in which the agent \textit{also} serves as the instruction-verification LLM: as the task is made easier for the instruction-verification model (see Section \ref{sec_randomworld_sub_env_sub_instructions}), an agent could still learn from tasks whose instructions it has verified itself. 

Under such an approach, the agent would train until it has mastered the data, then be used to verify instructions for more difficult data\textemdash on which it would then be further trained on\textemdash and so on. We leave the implementation of this approach to future work.

\section*{Acknowledgments}

We gratefully acknowledge the stimulating research environment of the GRK 2853/1 ``Neuroexplicit Models of Language, Vision, and Action'', funded by the Deutsche Forschungsgemeinschaft (DFG; German Research Foundation) under project number 471607914.

\bibliography{anthology,custom}

\begin{thebibliography}{34}
\providecommand{\natexlab}[1]{#1}

\bibitem[{Basu et~al.(2024)Basu, Abdelaziz, Kate, Agarwal, Crouse, Rizk, Bradford, Munawar, Kumaravel, Goyal et~al.}]{basu2024nestful}
Kinjal Basu, Ibrahim Abdelaziz, Kiran Kate, Mayank Agarwal, Maxwell Crouse, Yara Rizk, Kelsey Bradford, Asim Munawar, Sadhana Kumaravel, Saurabh Goyal, et~al. 2024.
\newblock Nestful: A benchmark for evaluating llms on nested sequences of api calls.
\newblock \emph{arXiv preprint arXiv:2409.03797}.

\bibitem[{Cai et~al.(2025)Cai, Li, Wang, Zhu, Shen, Li, and Chua}]{cai2025large}
Hongru Cai, Yongqi Li, Wenjie Wang, Fengbin Zhu, Xiaoyu Shen, Wenjie Li, and Tat-Seng Chua. 2025.
\newblock Large language models empowered personalized web agents.
\newblock In \emph{Proceedings of the ACM on Web Conference 2025}, pages 198--215.

\bibitem[{Chu et~al.(2025)Chu, Zhai, Yang, Tong, Xie, Schuurmans, Le, Levine, and Ma}]{chu2025sft}
Tianzhe Chu, Yuexiang Zhai, Jihan Yang, Shengbang Tong, Saining Xie, Dale Schuurmans, Quoc~V Le, Sergey Levine, and Yi~Ma. 2025.
\newblock Sft memorizes, rl generalizes: A comparative study of foundation model post-training.
\newblock \emph{arXiv preprint arXiv:2501.17161}.

\bibitem[{Davidson et~al.(2025)Davidson, Seguin, Bacis, Ilharco, and Harkous}]{davidson2025orchestrating}
Tim~R Davidson, Benoit Seguin, Enrico Bacis, Cesar Ilharco, and Hamza Harkous. 2025.
\newblock Orchestrating synthetic data with reasoning.
\newblock In \emph{ICLR 2025 Workshop on Synth Data}.

\bibitem[{Feng et~al.(2025)Feng, Huang, Qu, Zhang, Qin, Zhong, Jiang, Chi, and Zhong}]{feng2025retool}
Jiazhan Feng, Shijue Huang, Xingwei Qu, Ge~Zhang, Yujia Qin, Baoquan Zhong, Chengquan Jiang, Jinxin Chi, and Wanjun Zhong. 2025.
\newblock Retool: Reinforcement learning for strategic tool use in llms.
\newblock \emph{arXiv preprint arXiv:2504.11536}.

\bibitem[{Guo et~al.(2025)Guo, Yang, Zhang, Song, Zhang, Xu, Zhu, Ma, Wang, Bi et~al.}]{guo2025deepseek}
Daya Guo, Dejian Yang, Haowei Zhang, Junxiao Song, Ruoyu Zhang, Runxin Xu, Qihao Zhu, Shirong Ma, Peiyi Wang, Xiao Bi, et~al. 2025.
\newblock Deepseek-r1: Incentivizing reasoning capability in llms via reinforcement learning.
\newblock \emph{arXiv preprint arXiv:2501.12948}.

\bibitem[{Hu et~al.(2022)Hu, Shen, Wallis, Allen-Zhu, Li, Wang, Wang, Chen et~al.}]{hu2022lora}
Edward~J Hu, Yelong Shen, Phillip Wallis, Zeyuan Allen-Zhu, Yuanzhi Li, Shean Wang, Lu~Wang, Weizhu Chen, et~al. 2022.
\newblock Lora: Low-rank adaptation of large language models.
\newblock \emph{International Conference on Learning Representations}, 1(2):3.

\bibitem[{Hu et~al.(2024)Hu, Zhao, Xu, Sun, Lou, Lin, Luo, and Rajmohan}]{hu2024agentgen}
Mengkang Hu, Pu~Zhao, Can Xu, Qingfeng Sun, Jianguang Lou, Qingwei Lin, Ping Luo, and Saravan Rajmohan. 2024.
\newblock Agentgen: Enhancing planning abilities for large language model based agent via environment and task generation.
\newblock \emph{arXiv preprint arXiv:2408.00764}.

\bibitem[{Huang et~al.(2024)Huang, Zhong, Lu, Zhu, Gao, Liu, Hou, Zeng, Wang, Shang, Jiang, Xu, and Liu}]{huang-etal-2024-planning-creation}
Shijue Huang, Wanjun Zhong, Jianqiao Lu, Qi~Zhu, Jiahui Gao, Weiwen Liu, Yutai Hou, Xingshan Zeng, Yasheng Wang, Lifeng Shang, Xin Jiang, Ruifeng Xu, and Qun Liu. 2024.
\newblock \href {https://doi.org/10.18653/v1/2024.findings-acl.259} {Planning, creation, usage: Benchmarking {LLM}s for comprehensive tool utilization in real-world complex scenarios}.
\newblock In \emph{Findings of the Association for Computational Linguistics: ACL 2024}, pages 4363--4400, Bangkok, Thailand. Association for Computational Linguistics.

\bibitem[{Hurst et~al.(2024)Hurst, Lerer, Goucher, Perelman, Ramesh, Clark, Ostrow, Welihinda, Hayes, Radford et~al.}]{hurst2024gpt}
Aaron Hurst, Adam Lerer, Adam~P Goucher, Adam Perelman, Aditya Ramesh, Aidan Clark, AJ~Ostrow, Akila Welihinda, Alan Hayes, Alec Radford, et~al. 2024.
\newblock Gpt-4o system card.
\newblock \emph{arXiv preprint arXiv:2410.21276}.

\bibitem[{Li et~al.(2023)Li, Zhao, Yu, Song, Li, Yu, Li, Huang, and Li}]{li-etal-2023-api}
Minghao Li, Yingxiu Zhao, Bowen Yu, Feifan Song, Hangyu Li, Haiyang Yu, Zhoujun Li, Fei Huang, and Yongbin Li. 2023.
\newblock \href {https://doi.org/10.18653/v1/2023.emnlp-main.187} {{API}-bank: A comprehensive benchmark for tool-augmented {LLM}s}.
\newblock In \emph{Proceedings of the 2023 Conference on Empirical Methods in Natural Language Processing}, pages 3102--3116, Singapore. Association for Computational Linguistics.

\bibitem[{Li et~al.(2025)Li, Dong, Jin, Zhang, Zhou, Zhu, Zhang, and Dou}]{li2025search}
Xiaoxi Li, Guanting Dong, Jiajie Jin, Yuyao Zhang, Yujia Zhou, Yutao Zhu, Peitian Zhang, and Zhicheng Dou. 2025.
\newblock Search-o1: Agentic search-enhanced large reasoning models.
\newblock \emph{arXiv preprint arXiv:2501.05366}.

\bibitem[{Lin et~al.(2025)Lin, Wen, Peng, Nie, Liao, Wang, Mo, Zhou, Cheng, Zhao, Wang, and Weinan}]{lin2025robust}
Qiqiang Lin, Muning Wen, Qiuying Peng, Quanyu Nie, Junwei Liao, Jun Wang, Xiaoyun Mo, Jiamu Zhou, Cheng Cheng, Yin Zhao, Jun Wang, and Zhang Weinan. 2025.
\newblock Robust function-calling for on-device language model via function masking.
\newblock In \emph{The Thirteenth International Conference on Learning Representations}.

\bibitem[{Liu et~al.(2024{\natexlab{a}})Liu, Huang, Zeng, Hao, Yu, Li, Wang, Gan, Liu, Yu et~al.}]{liu2024toolace}
Weiwen Liu, Xu~Huang, Xingshan Zeng, Xinlong Hao, Shuai Yu, Dexun Li, Shuai Wang, Weinan Gan, Zhengying Liu, Yuanqing Yu, et~al. 2024{\natexlab{a}}.
\newblock Toolace: Winning the points of llm function calling.
\newblock \emph{arXiv preprint arXiv:2409.00920}.

\bibitem[{Liu et~al.(2024{\natexlab{b}})Liu, Hoang, Zhang, Zhu, Lan, Tan, Yao, Liu, Feng, RN et~al.}]{liu2024apigen}
Zuxin Liu, Thai Hoang, Jianguo Zhang, Ming Zhu, Tian Lan, Juntao Tan, Weiran Yao, Zhiwei Liu, Yihao Feng, Rithesh RN, et~al. 2024{\natexlab{b}}.
\newblock Apigen: Automated pipeline for generating verifiable and diverse function-calling datasets.
\newblock \emph{Advances in Neural Information Processing Systems}, 37:54463--54482.

\bibitem[{Matthews et~al.(2024)Matthews, Beukman, Lu, and Foerster}]{matthews2024kinetix}
Michael Matthews, Michael Beukman, Chris Lu, and Jakob Foerster. 2024.
\newblock Kinetix: Investigating the training of general agents through open-ended physics-based control tasks.
\newblock \emph{arXiv preprint arXiv:2410.23208}.

\bibitem[{McDermott et~al.(1998)McDermott, Ghallab, Howe, Knoblock, Ram, Veloso, Weld, and Wilkins}]{McDermott1998PDDLthePD}
Drew McDermott, Malik Ghallab, Adele~E. Howe, Craig~A. Knoblock, Ashwin Ram, Manuela~M. Veloso, Daniel~S. Weld, and David~E. Wilkins. 1998.
\newblock \href {https://api.semanticscholar.org/CorpusID:59656859} {Pddl-the planning domain definition language}.

\bibitem[{Patil et~al.(2024)Patil, Zhang, Wang, and Gonzalez}]{patil2024gorilla}
Shishir~G Patil, Tianjun Zhang, Xin Wang, and Joseph~E Gonzalez. 2024.
\newblock Gorilla: Large language model connected with massive apis.
\newblock \emph{Advances in Neural Information Processing Systems}, 37:126544--126565.

\bibitem[{Qi et~al.(2025)Qi, Liu, Iong, Lai, Xueqiao, Sun, Yang, Yang, Yao, Xu, Tang, and Dong}]{qi2025webrl}
Zehan Qi, Xiao Liu, Iat~Long Iong, Hanyu Lai, Sun Xueqiao, Jiadai Sun, Xinyue Yang, Yu~Yang, Shuntian Yao, Wei Xu, Jie Tang, and Yuxiao Dong. 2025.
\newblock Web{RL}: Training {LLM} web agents via self-evolving online curriculum reinforcement learning.
\newblock In \emph{The Thirteenth International Conference on Learning Representations}.

\bibitem[{Qian et~al.(2025)Qian, Acikgoz, He, Wang, Chen, Hakkani-T{\"u}r, Tur, and Ji}]{qian2025toolrl}
Cheng Qian, Emre~Can Acikgoz, Qi~He, Hongru Wang, Xiusi Chen, Dilek Hakkani-T{\"u}r, Gokhan Tur, and Heng Ji. 2025.
\newblock Toolrl: Reward is all tool learning needs.
\newblock \emph{arXiv preprint arXiv:2504.13958}.

\bibitem[{Qin et~al.(2024)Qin, Liang, Ye, Zhu, Yan, Lu, Lin, Cong, Tang, Qian, Zhao, Hong, Tian, Xie, Zhou, Gerstein, Li, Liu, and Sun}]{qin2024toolllm}
Yujia Qin, Shihao Liang, Yining Ye, Kunlun Zhu, Lan Yan, Yaxi Lu, Yankai Lin, Xin Cong, Xiangru Tang, Bill Qian, Sihan Zhao, Lauren Hong, Runchu Tian, Ruobing Xie, Jie Zhou, Mark Gerstein, Dahai Li, Zhiyuan Liu, and Maosong Sun. 2024.
\newblock Tool{LLM}: Facilitating large language models to master 16000+ real-world {API}s.
\newblock In \emph{The Twelfth International Conference on Learning Representations}.

\bibitem[{Rafailov et~al.(2023)Rafailov, Sharma, Mitchell, Manning, Ermon, and Finn}]{rafailov2023direct}
Rafael Rafailov, Archit Sharma, Eric Mitchell, Christopher~D Manning, Stefano Ermon, and Chelsea Finn. 2023.
\newblock Direct preference optimization: Your language model is secretly a reward model.
\newblock \emph{Advances in Neural Information Processing Systems}, 36:53728--53741.

\bibitem[{Shao et~al.(2024)Shao, Wang, Zhu, Xu, Song, Bi, Zhang, Zhang, Li, Wu et~al.}]{shao2024deepseekmath}
Zhihong Shao, Peiyi Wang, Qihao Zhu, Runxin Xu, Junxiao Song, Xiao Bi, Haowei Zhang, Mingchuan Zhang, YK~Li, Y~Wu, et~al. 2024.
\newblock Deepseekmath: Pushing the limits of mathematical reasoning in open language models.
\newblock \emph{arXiv preprint arXiv:2402.03300}.

\bibitem[{Trivedi et~al.(2024)Trivedi, Khot, Hartmann, Manku, Dong, Li, Gupta, Sabharwal, and Balasubramanian}]{trivedi-etal-2024-appworld}
Harsh Trivedi, Tushar Khot, Mareike Hartmann, Ruskin Manku, Vinty Dong, Edward Li, Shashank Gupta, Ashish Sabharwal, and Niranjan Balasubramanian. 2024.
\newblock \href {https://doi.org/10.18653/v1/2024.acl-long.850} {{A}pp{W}orld: A controllable world of apps and people for benchmarking interactive coding agents}.
\newblock In \emph{Proceedings of the 62nd Annual Meeting of the Association for Computational Linguistics (Volume 1: Long Papers)}, pages 16022--16076, Bangkok, Thailand. Association for Computational Linguistics.

\bibitem[{von Werra et~al.(2020)von Werra, Belkada, Tunstall, Beeching, Thrush, Lambert, Huang, Rasul, and Gallouédec}]{vonwerra2022trl}
Leandro von Werra, Younes Belkada, Lewis Tunstall, Edward Beeching, Tristan Thrush, Nathan Lambert, Shengyi Huang, Kashif Rasul, and Quentin Gallouédec. 2020.
\newblock {TRL: Transformer Reinforcement Learning}.
\newblock \url{https://github.com/huggingface/trl}.

\bibitem[{Wang et~al.(2019)Wang, Pruksachatkun, Nangia, Singh, Michael, Hill, Levy, and Bowman}]{wang2019superglue}
Alex Wang, Yada Pruksachatkun, Nikita Nangia, Amanpreet Singh, Julian Michael, Felix Hill, Omer Levy, and Samuel Bowman. 2019.
\newblock Superglue: A stickier benchmark for general-purpose language understanding systems.
\newblock \emph{Advances in Neural Information Processing Systems}, 32.

\bibitem[{Yan et~al.(2024)Yan, Mao, Ji, Zhang, Patil, Stoica, and Gonzalez}]{berkeley-function-calling-leaderboard}
Fanjia Yan, Huanzhi Mao, Cheng-Jie Ji, Tianjun Zhang, Shishir~G. Patil, Ion Stoica, and Joseph~E. Gonzalez. 2024.
\newblock Berkeley function calling leaderboard.

\bibitem[{Yang et~al.(2024)Yang, Yang, Zhang, Hui, Zheng, Yu, Li, Liu, Huang, Wei et~al.}]{yang2024qwen2}
An~Yang, Baosong Yang, Beichen Zhang, Binyuan Hui, Bo~Zheng, Bowen Yu, Chengyuan Li, Dayiheng Liu, Fei Huang, Haoran Wei, et~al. 2024.
\newblock Qwen2.5 technical report.
\newblock \emph{arXiv preprint arXiv:2412.15115}.

\bibitem[{Yao et~al.(2022)Yao, Chen, Yang, and Narasimhan}]{yao2022webshop}
Shunyu Yao, Howard Chen, John Yang, and Karthik Narasimhan. 2022.
\newblock Webshop: Towards scalable real-world web interaction with grounded language agents.
\newblock \emph{Advances in Neural Information Processing Systems}, 35:20744--20757.

\bibitem[{Yao et~al.(2023)Yao, Zhao, Yu, Du, Shafran, Narasimhan, and Cao}]{yao2023react}
Shunyu Yao, Jeffrey Zhao, Dian Yu, Nan Du, Izhak Shafran, Karthik Narasimhan, and Yuan Cao. 2023.
\newblock React: Synergizing reasoning and acting in language models.
\newblock In \emph{International Conference on Learning Representations (ICLR)}.

\bibitem[{Zhang et~al.(2024)Zhang, Lan, Zhu, Liu, Hoang, Kokane, Yao, Tan, Prabhakar, Chen et~al.}]{zhang2024xlam}
Jianguo Zhang, Tian Lan, Ming Zhu, Zuxin Liu, Thai Hoang, Shirley Kokane, Weiran Yao, Juntao Tan, Akshara Prabhakar, Haolin Chen, et~al. 2024.
\newblock xlam: A family of large action models to empower ai agent systems.
\newblock \emph{arXiv preprint arXiv:2409.03215}.

\bibitem[{Zheng et~al.(2025)Zheng, Fu, Hu, Cai, Ye, Lu, and Liu}]{zheng2025deepresearcher}
Yuxiang Zheng, Dayuan Fu, Xiangkun Hu, Xiaojie Cai, Lyumanshan Ye, Pengrui Lu, and Pengfei Liu. 2025.
\newblock Deepresearcher: Scaling deep research via reinforcement learning in real-world environments.
\newblock \emph{arXiv preprint arXiv:2504.03160}.

\bibitem[{Zheng et~al.(2024)Zheng, Sun, Qiu, Ru, Jiayang, Li, Lin, Wang, Luo, Pan, Xu, Min, Zhang, Wang, Li, and Liu}]{zheng-etal-2024-openresearcher}
Yuxiang Zheng, Shichao Sun, Lin Qiu, Dongyu Ru, Cheng Jiayang, Xuefeng Li, Jifan Lin, Binjie Wang, Yun Luo, Renjie Pan, Yang Xu, Qingkai Min, Zizhao Zhang, Yiwen Wang, Wenjie Li, and Pengfei Liu. 2024.
\newblock \href {https://doi.org/10.18653/v1/2024.emnlp-demo.22} {{O}pen{R}esearcher: Unleashing {AI} for accelerated scientific research}.
\newblock In \emph{Proceedings of the 2024 Conference on Empirical Methods in Natural Language Processing: System Demonstrations}, pages 209--218, Miami, Florida, USA. Association for Computational Linguistics.

\bibitem[{Zhuang et~al.(2023)Zhuang, Yu, Wang, Sun, and Zhang}]{zhuang2023toolqa}
Yuchen Zhuang, Yue Yu, Kuan Wang, Haotian Sun, and Chao Zhang. 2023.
\newblock Toolqa: A dataset for llm question answering with external tools.
\newblock \emph{Advances in Neural Information Processing Systems}, 36:50117--50143.

\end{thebibliography}

\appendix

\begin{figure*}[]
\centering
\includegraphics[width=\linewidth]{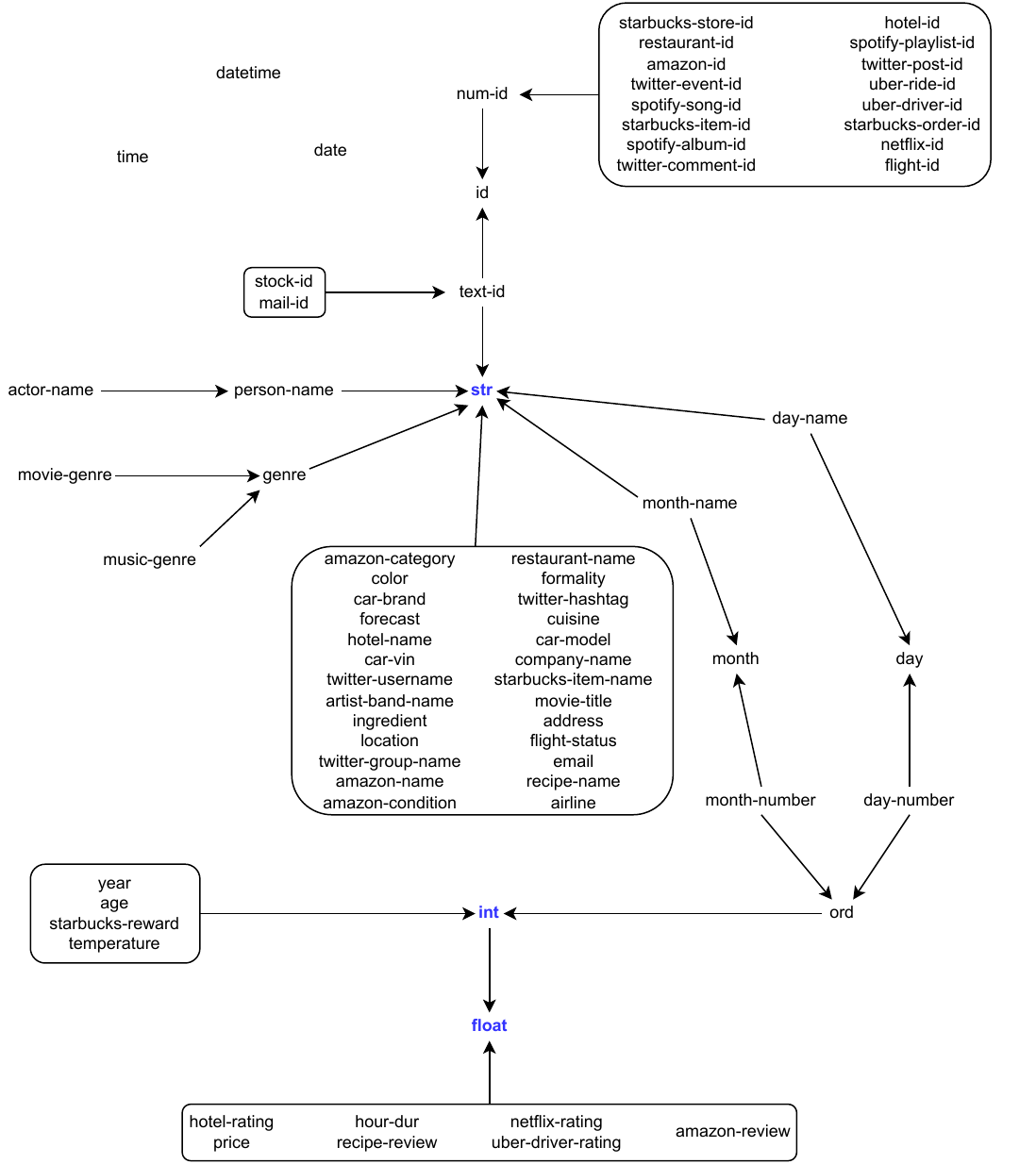}
\caption{The RandomWorld type hierarchy: actual Python types are bolded and in blue\textemdash all other nodes denote custom RandomWorld types. Edges $A\rightarrow B$ indicates that $A$ is a subtype of $B$ ($A\leq B$). For the sake of representational simplicity, an edge $\square\rightarrow A$ indicates that all types within the box are subtypes of $A$, without any subtype relations between types within the box: for example, $\textit{stock-id},\textit{mail-id}\leq\textit{text-id}$.}
\label{fig_type_hierarchy}
\end{figure*}

\begin{table*}[]
\centering
\scalebox{0.85}{\begin{tabular}{l|ll}
\textbf{Name} & \textbf{Description} & \textbf{Example(s)} \\
\hline
\textit{actor-name} & name of an actor & "Leonardo DiCaprio", "Meryl Streep" \\
\textit{address} & location of a building or plot of land & "123 Maple Street, Springfield, IL 62701, USA" \\
\textit{age} & age in years & 13, 98 \\
\textit{airline} & name of an airline & "American Airlines", "Delta Air Lines" \\
\textit{amazon-category} & shopping category on Amazon & "Books", "Electronics" \\
\textit{amazon-condition} & condition of an Amazon item & "New", "Used, Like New" \\
\textit{amazon-id} & numerical ID of an Amazon item & 430680496270, 478108090872052 \\
\textit{amazon-name} & name of an Amazon item & "Toilet Paper", "Paper Towels" \\
\textit{amazon-review} & rating of an Amazon item & 4.8, 0.2 \\
\textit{artist-band-name} & name of a (music) band & "The Beatles", "Rolling Stones" \\
\textit{car-brand} & name of a car manufacturer & "Toyota", "Ford" \\
\textit{car-model} & name of a car model & "LaCrosse", "Phantom" \\
\textit{car-vin} & vehicle identification number & "gjuqfykjulqsitv7r", "lvc3qd874emg411nl" \\
\textit{color} & color & "Red", "Green" \\
\textit{company-name} & name of a company & "Apple", "Microsoft" \\
\textit{cuisine} & name of a category of food & "Italian", "Chinese" \\
\textit{date} & date & "17/8/1103", "20/5/183" \\
\textit{datetime} & date and time & "05:01 4/10/1302", "11:33 9/2/100" \\
\textit{day-name} & name of a day of the week & "Monday", "Tuesday" \\
\textit{day-number} & calendar day number & 1, 2 \\
\textit{email} & content of an email message & "Dear Mr. Johnson, ...", "Hi Sarah, ..." \\
\textit{flight-id} & numerical ID of a commercial flight & 222101709170966, 9765628923380 \\
\textit{flight-status} & status of a flight & "On Time", "Cancelled" \\
\textit{forecast} & weather forecast & "Clear", "Partly Cloudy" \\
\textit{formality} & tone of a text & "formal", "semi-formal" \\
\textit{hotel-id} & numerical ID of a hotel & 4496768012, 3155376944 \\
\textit{hotel-name} & name of a hotel & "The Grand Magnolia", "Skyline Retreat" \\
\textit{hotel-rating} & rating of a hotel & 1.5, 1.0 \\
\textit{hour-dur} & length of time (in hours) & 1.2, 1.0 \\
\textit{ingredient} & name of an ingredient & "Garlic", "Onion" \\
\textit{location} & geographic location & "New York", "Los Angeles" \\
\textit{mail-id} & email address & "i8njw1s@oj7y.ca", "k@dy851wvil2vy4z7by.com" \\
\textit{month-name} & name of a month & "January", "February" \\
\textit{month-number} & calendar month number (1-12) & 1, 2 \\
\textit{movie-genre} & genre of a movie & "Action", "Adventure" \\
\textit{movie-title} & title of a movie & "The Shawshank Redemption", "The Godfather" \\
\textit{music-genre} & genre of a song or album & "Rock", "Pop" \\
\textit{netflix-id} & numerical ID of a movie on Netflix & 2737985392929, 1724771805351 \\
\textit{netflix-rating} & rating of a movie on Netflix & 1.8, 0.0 \\
\textit{person-name} & name of a person & "John Doe", "Sarah Smith" \\
\textit{price} & cost of an item & 627.49, 4545.56 \\
\textit{recipe-name} & name of a recipe & "Spaghetti Carbonara", "Chicken Alfredo" \\
\textit{recipe-review} & rating of a recipe & 1.4, 0.4 \\
\textit{restaurant-id} & numerical ID of a restaurant & 2354517290620, 82682880027029 \\
\textit{restaurant-name} & name of a restaurant & "The Golden Spoon", "Bella Cucina" \\
\textit{spotify-album-id} & numerical ID of a Spotify album & 54529623464, 6035921619 \\
\textit{spotify-playlist-id} & numerical ID of a Spotify playlist & 565019246380, 339078364981 \\
\textit{spotify-song-id} & numerical ID of a song on Spotify & 74587698010, 16030267814 \\
\textit{starbucks-item-id} & numerical ID of a Starbucks product & 2847706651, 61785093490348 \\
\textit{starbucks-item-name} & name of a Starbucks product & "Caramel Macchiato", "Caffè Latte"
\end{tabular}}
\caption{RandomWorld custom type names, descriptions, and 1-2 example instances (depending on length), sampled from their respective generators (continued in Table \ref{table_type_exs2}). Supertypes are not displayed, as their generators and recognizers are inherited from their constituent subtypes.}
\label{table_type_exs1}
\end{table*}

\begin{table*}[]
\centering
\scalebox{0.85}{\begin{tabular}{l|ll}
\textbf{Name} & \textbf{Description} & \textbf{Example(s)} \\
\hline
\textit{starbucks-order-id} & numerical ID of a Starbucks order & 4251443807182, 25283848032 \\
\textit{starbucks-reward} & number of Starbucks reward points & 430, 257 \\
\textit{starbucks-store-id} & numerical ID of a Starbucks store & 128977859410, 915762041007 \\
\textit{stock-id} & stock ticker symbol of a company & "WPHL", "L" \\
\textit{temperature} & temperature & 21.5, 37.0 \\
\textit{time} & time & "23:37", "17:47" \\
\textit{twitter-comment-id} & numerical ID of a comment on a Twitter post & 8458186204222, 129076645933 \\
\textit{twitter-event-id} & numerical ID of a Twitter event & 537044647554022, 51090184304 \\
\textit{twitter-group-name} & name of a Twitter group & "TechTalks", "FoodieFriends" \\
\textit{twitter-hashtag} & Twitter hashtag & "\#FollowFriday", "\#TechNews" \\
\textit{twitter-post-id} & numerical ID of a Twitter post & 34268389893, 73626538108 \\
\textit{twitter-username} & username of a Twitter account & "JohnDoe", "SarahSmith1" \\
\textit{uber-driver-id} & numerical ID of an Uber driver & 617303431222, 6236243225 \\
\textit{uber-driver-rating} & rating of an Uber driver & 4.2, 2.8 \\
\textit{uber-ride-id} & numerical ID of an Uber ride & 6346853814, 90798212714679 \\
\textit{year} & year & 1841, 1988
\end{tabular}}
\caption{Table \ref{table_type_exs1} continued.}
\label{table_type_exs2}
\end{table*}

\begin{table*}[]
\centering
\scalebox{0.94}{\begin{tabular}{l|l}
\textbf{Name: Signature} & \textbf{Description} \\
\hline
$\textit{actor-movie}\colon\textit{actor-name}\to\textit{list}(\textit{movie-title})$ & retrieves the names of movies in which a \\
 & particular actor plays \\
 & \\
$\textit{age-movie}\colon\textit{union}(\textit{movie-title},\textit{netflix-id})\to\textit{age}$ & retrieves the age for which the input movie is \\
 & appropriate \\
 & \\
$\textit{daily-ingredient-specials}\colon$ & returns a dictionary of daily special ingredients \\
$\textit{day-name}\to\textit{dict}(\textit{ingredient},\textit{restaurant-name})$ & and their associated restaurants \\
 & \\
$\textit{dining-time-matcher}\colon\textit{age}\to(\textit{time}\times\textit{restaurant-name})$ & suggests dining time and restaurant based on \\
 & age preferences \\
 & \\
$\textit{frequent-day-finder}\colon$ & returns the most common day from the \\
$\textit{dict}(\textit{restaurant-id},\textit{day-name})\to\textit{day-name}$ & restaurant-day mapping \\
 & \\
$\textit{holiday-checker}\colon\textit{location}\to\textit{date}$ & returns the most recent public holiday at a \\
 & location \\
 & \\
$\textit{hq-locator}\colon\textit{company-name}\to\textit{location}$ & returns the headquarters location of the input \\
 & company \\
 & \\
$\textit{movie-len}\colon\textit{hour-dur}\times\textit{hour-dur}\to\textit{list}(\textit{movie-title})$ & retrieves movies that last between the specified \\
 & time range \\
 & \\
$\textit{recipe-suggester}\colon\textit{day}\to\textit{recipe-name}$ & suggests a recipe based on the given day \\
 & \\
$\textit{starbucks-locator}\colon\textit{location}\to\textit{starbucks-store-id}$ & returns the nearest Starbucks store ID for a given \\
 & location \\
 & \\
$\textit{stock-price}\colon\textit{stock-id}\times\textit{date}\to\textit{price}$ & retrieves the price for a given stock ticker symbol \\
 & on the specified date \\
 & \\
$\textit{stock-ticker}\colon\textit{company-name}\to\textit{stock-id}$ & returns the stock ticker symbol associated with a \\
 & given company name \\
 
\end{tabular}}
\caption{Example tool names/signatures and descriptions generated by RandomWorld (see Section \ref{sec_randomworld_sub_tool}).}
\label{table_tool_exs}
\end{table*}

\newpage

\hspace{0mm}

\newpage

\section{Type System}
\label{app_types}

The majority of the RandomWorld type system is described in detail in Section \ref{sec_randomworld_sub_types}, Figure \ref{fig_type_hierarchy}, and Tables \ref{table_type_exs1}/\ref{table_type_exs2}: this section is primarily dedicated a more detailed discussion of the constructed types discussed in Section \ref{sec_randomworld_sub_types}.

\subsection{Subtype Relations}
\label{app_types_sub_subtypes}

Subtype relations between atomic types (i.e.\hspace{1mm}those given in Figure \ref{fig_type_hierarchy}) are defined manually; subtype relations between constructed types are defined recursively in terms of their component types, where the recursion is terminated by the atomic types.

Subtype relations between list types are defined as in Equation \ref{eq_list_subtype}.

\begin{equation}
\label{eq_list_subtype}
\textit{list}(A)\leq B:=\exists C(B=\textit{list}(C)\land A\leq C)
\end{equation}

This is to say that $\textit{list}(A)$ is a subtype of $B$ if $B$ is of the form $\textit{list}(C)$ and $A$ is a subtype of $C$: list types cannot be supertypes or subtypes of non-list types.

Subtype relations between dictionary types are defined as in Equation \ref{eq_dict_subtype}.

\begin{equation}
\label{eq_dict_subtype}
\begin{split}
    \textit{dict}&(A,B)\leq C:=\\\exists &X,Y(B=\textit{dict}(X,Y)\land A\geq X\land B\leq Y)
\end{split}
\end{equation}

This is to say that $\textit{dict}(A,B)$ is a subtype of $C$ if $C$ is of the form $\textit{dict}(X,Y)$, $B\leq Y$, and $X\leq A$: as mappings, dictionary types are contravariant in their first argument. As with list types, dictionary types cannot be supertypes or subtypes of non-dictionary types.

Sub-/super-type relations for union types are defined as in Equation \ref{eq_union_subtype}.

\begin{subequations}
\label{eq_union_subtype}
\begin{align}
&\textit{union}(A,B)\leq C:=A\leq C\land B\leq C \\
&C\leq\textit{union}(A,B):=C\leq A\lor C\leq B
\end{align}
\end{subequations}

Note that union types are \textit{not} discriminated unions (coproducts)\textemdash for example:

\begin{equation*}
\begin{split}
    \textit{union}(&A,\textit{union}(B,C))=\textit{union}(\textit{union}(A,B),C) \\
    &=\textit{union}(\textit{union}(A,B),\textit{union}(B,C))
\end{split}
\end{equation*}


\subsection{Generators}
\label{app_types_sub_generators}

Type generators for atomic types are defined manually; generators for constructed types are defined recursively in terms of their components. 

Generators $G_{\textit{union}(A,B)}$ for union types $\textit{union}(A,B)$ simply randomly sample one of their arguments $X\sim\{A,B\}$, then sample an output from the generator $G_X$.

Generators $G_{\textit{list}(A)}$ for list types $\textit{list}(A)$ first sample a length $\ell$ (within a pre-defined range), then construct a list by sampling $\ell$ outputs from $G_A$.

Similarly, generators $G_{\textit{dict}(A,B)}$ for dictionary types $\textit{dict}(A,B)$ sample a length $\ell$, then construct a dictionary by sampling $\ell$ pairs $(a,b)$, with $a\sim G_A$ and $b\sim G_B$.

\subsection{Recognizers}
\label{app_types_sub_recognizers}

As discussed in Section \ref{sec_randomworld_sub_types}, type recognizers are boolean-valued functions $R_A\colon \text{Any}\to\{0,1\}$, such that $R_A(x)=1$ if $x$ is of type $A$, and $R_A(x)=0$ otherwise.

As with generators, type recognizers for atomic types are defined manually\textemdash e.g.\hspace{1mm}through regular expressions, set membership checks, etc., depending on the type\textemdash while recognizers $R_A$ for constructed types $A$ are defined recursively in terms of their components, as in Equation \ref{eq_recognizers}, where $I(x,A):=\textit{isinstance}(x,A)$ for a Python type $A$.

\begin{subequations}
\label{eq_recognizers}
\begin{align}
&R_{\textit{union}(A,B)}(x):=R_A(x)\lor R_B(x) \\
&R_{\textit{list}(A)}(x):=I(x,\textit{list})\land\underset{i=0}{\overset{|x|-1}{\bigwedge}}R_A(x_i) \\
\begin{split}
    R&_{\textit{dict}(A,B)}(x):=\\ 
    &I(x,\textit{dict})\land\underset{(a,b)\in \textit{items}(x)}{\bigwedge}\left(R_A(a)\land R_B(b)\right)
\end{split}
\end{align}
\end{subequations}

\subsection{Supertypes}
\label{app_types_sub_supertypes}

As discussed in Section \ref{sec_randomworld_sub_types}, supertypes in the hierarchy in Figure \ref{fig_type_hierarchy} inherit subtype relations, recongizers, and generators from their subtypes\textemdash with the exception of Python types, whose recognizers are defined by the built-in \textit{isinstance} function. 

For a given type $A$ with subtypes $B_1,\dots, B_n$, we implement subtype relation and recongizer/generator inheritance by implicitly defining $A$ as $\textit{union}(B_1,\dots,B_n)$.

\section{Example Synthetic Tasks}
\label{app_trjs}

\subsection{GRPO (Llama-RW-GRPO)}
\label{app_trjs_sub_grpo}

Agent-generated tokens are \hl{highlighted}, and correct/incorrect submissions are indicated with a green check/red X (\gcheck/\rex), respectively.

\hspace{0mm}
\\

\hspace*{-3.85mm}\fbox{%
    \parbox{74mm}{\tiny
        The tools at your disposal: \\
        netflix\_movie-age: retrieves movie titles appropriate for children at or under the specified age \\
        netflix\_actor-movie: retrieves the name of the actors that play in a given movie \\ netflix\_age-movie: retrieves the age for which the input movie is appropriate \\
        netflix\_movie-long: retrieves a movie that lasts more than n hours \\
        mail-formality: retrieves emails in the user's inbox by the formality of the greeting \\
        netflix\_genre-movie: retrieves the genre of a specified movie \\
        netflix\_list: retrieves the ID of the nth-most recently-added movie to the user's list \\ 
        netflix\_movie-genre: retrieves a movie recommendation based on the input genre \\
        spotify\_recent: retrieves the ID of the user's nth-most recently listened-to song \\ spotify\_top: retrieves the ID of the nth-most listened-to song in the user's playlist \\\\
        Find a movie longer than 1.2 hours, get its genre, recommend another movie of the same genre, determine the appropriate age for that movie, and then find a list of movies suitable for that age. My Netflix username is w1j6w9d294ubn2jutf. \\\\\\
        \hl{First, I need to log in to Netflix.} \\ \hl{<request><password-manager>`netflix',`wlj6w9d294ubn2jutf'<call>}`rbn3atn50ofc' \\ 
        <response> \\\hl{<request><netflix\_login>'w1j6w9d294ubn2jutf','rbn3atn50ofc'<call>} SUCCESS\\ 
        <response> \\\\
        \hl{Now, I need to find a movie longer than 1.2 hours.} \\
        \hl{<request><netflix\_movie-long>1.2<call>}`Compromises'<response> \\\\
        Now, I need to get the genre of the movie ``Compromises''. \\
        \hl{<request><netflix\_genre-movie>``Compromises''}<call>`Drama'<response> \\\\
        \hl{Now, I need to recommend another movie of the same genre.} \\
        \hl{<request><netflix\_movie-genre>`Drama'<call>}`Fight Club'<response> \\\\
        \hl{Now, I need to determine the appropriate age for the movie ``Fight Club''.} \\ \hl{<request><netflix\_age-movie>``Fight Club''<call>}18<response> \\\\
        \hl{Now, I need to find a list of movies suitable for children at or under the age of 18.} \\
        \hl{<request><netflix\_movie-age>18<call>}[`The Dark Knight', `Pulp Fiction', `Up', `Mad Max: Fury Road', `Toy Story', `Moana', `Zootopia', `Frozen', `The Lion King', `Fight Club', `The Incredibles', `Guardians of the Galaxy']<response> \\\\
        \hl{Result=[`The Dark Knight', `Pulp Fiction', `Up', `Mad Max: Fury Road', `Toy Story', `Moana', `Zootopia', `Frozen', `The Lion King', `Fight Club', `The Incredibles', `Guardians of the Galaxy']<submit>} \gcheck
    }%
}

\hspace{0mm}
\\

\hspace*{-3.85mm}\fbox{%
    \parbox{74mm}{\tiny
        The tools at your disposal: \\
        day-to-name-converter: Converts day input to corresponding weekday name. \\
        comment-month-extractor: Extracts the month number from a Twitter comment's timestamp. \\ 
        comment-day-finder: Returns the day of the week a Twitter comment was posted. \\
        flight-arrival-time: Returns the scheduled arrival time for a given flight ID. \\ 
        recipe-rating-fetcher: Retrieves average user rating for a specified recipe. \\
        hashtag-comment-finder: Finds a comment ID from a post with the specified hashtag. \\\\
        Find the weekday name on which a comment with the hashtag `\#AdventureAwaits' was posted. \\\\\\
        \hl{<request><hashtag-comment-finder>````AdventureAwaits''''<call>}\\
        Tool error: invalid input format<error> \\\\
        \hl{<request><hashtag-comment-finder>````\#AdventureAwaits''''<call>}\\
        Tool error: invalid input format<error> \\\\
        \hl{<request><hashtag-comment-finder>``\#AdventureAwaits''<call>}97876083389 \\ <response> \\\\
        \hl{<request><comment-day-finder>97876083389<call>}`Tuesday'<response> \\\\
        \hl{Result=`Tuesday'}<submit> \gcheck
    }%
}

\hspace{0mm}
\\

\hspace*{-3.85mm}\fbox{%
    \parbox{74mm}{\tiny
        The tools at your disposal: \\
        item-price: retrieves all items available at a specified price \\
        cart-checker: retrieves the ID of the nth-most recently-added item to the user's cart \\
        min: returns the minimum of a list of values \\
        price-item: retrieves all prices available for a specified item \\ 
        corporate-color-decoder: Determines the primary brand color associated with a given company name. \\
        last-serve-time: Returns the last service time for a specified restaurant ID \\\\
        Find all items available at the minimum price of laptops. \\\\\\
        \hl{<request><price-item>calculator-min(price-item(`laptop'))<call>} \\
        Tool error: invalid input format<error> \\\\
        \hl{Result=Tool error: invalid input format<submit>} \rex
    }%
}


\subsection{SFT}
\label{app_trjs_sub_grpo}

Tokens that contribute to the loss are \hl{highlighted}, all other tokens are masked from the loss function. 

\hspace{0mm}
\\

\hspace*{-3.85mm}\fbox{%
    \parbox{74mm}{\tiny
        The tools at your disposal:\\daily-recipe-suggester: Suggests a recipe based on the given day.\\sub: subtracts the second argument from the first\\extract-year-from-datetime: Extracts the year from a given datetime input.\\airline-founding-date: Returns the founding date of the specified airline.\\hashtag-trend-day: Returns the peak trending day for a given Twitter hashtag.\\recipe-rating-fetcher: Retrieves average user rating for a specified recipe.\\tweet-timestamp-retriever: Fetches the date and time a Twitter post was created using its ID.\\restaurant-age-checker: Returns the number of years since the restaurant's establishment.\\historical-event-date: Returns a significant historical date for the specified location.\\cuisine-to-location: Suggests a city known for the specified cuisine.\\store-temp-checker: Returns current temperature at specified Starbucks store location.\\tweet-locator: Fetches a recent Twitter post ID from the specified location.\\next-day-finder: Returns the next day of the week for the given input day.\\int-to-weekday: Converts an integer to a corresponding day of the week, modulo 7.\\tweet-day-extractor: Returns the calendar day a tweet was posted.\\movie-soundtrack-matcher: Returns a Spotify playlist ID inspired by the given movie title.\\\\Find a recipe based on the weekday corresponding to the year a recent tweet was posted from a city known for `Afghan' cuisine, retrieve its rating, and subtract the year from the rating.\\\\\\\hl{<request><cuisine-to-location>`Afghan'<call>}'Madison'<response>\\\hl{<request><tweet-locator>`Madison'<call>}4811550653611<response>\\\hl{<request><tweet-timestamp-retriever>4811550653611<call>}`10:05 2/8/2041'\\<response>\\\hl{<request><extract-year-from-datetime>`10:05 2/8/2041'<call>}2011<response>\\\hl{<request><int-to-weekday>2011<call>}`Friday'<response>\\\hl{<request><daily-recipe-suggester>`Friday'<call>}`Cinnamon Rolls'<response>\\\hl{<request><recipe-rating-fetcher>`Cinnamon Rolls'<call>}2.4<response>\\\hl{<request><sub>2.4,2011<call>}-2008.6<response>\\\hl{Result=-2008.6<submit>}
    }%
}

\hspace{0mm}
\\

\hspace*{-3.85mm}\fbox{%
    \parbox{74mm}{\tiny
        The tools at your disposal:\\hotel-id-lookup: Retrieves the unique ID for a specified hotel name.\\uber-driver-lookup: Retrieves the driver ID associated with a specific Uber ride ID.\\get-driver-rating: Fetches the current rating for a specified Uber driver by their ID.\\tone-refiner: Adjusts input tone to a similar, nuanced tone.\\starbucks-item-suggester: Suggests a complementary Starbucks item based on the input item name.\\starbucks-calorie-match: Suggests a Starbucks item based on the calorie range represented by the float.\\\\Suggest a Starbucks item based on the calorie range that matches the rating of the driver for Uber ride ID 6728308325.\\\\\\\hl{<request><uber-driver-lookup>6728308325<call>}78997392675<response>\\\hl{<request><get-driver-rating>78997392675<call>}1.4<response>\\\hl{<request><starbucks-calorie-match>1.4<call>}`Pumpkin Spice Latte'<response>\\\hl{Result=`Pumpkin Spice Latte'<submit>}
    }%
}

\hspace{0mm}
\\

\hspace*{-3.85mm}\fbox{%
    \parbox{74mm}{\tiny
        The tools at your disposal:\\spotify-album-release-date: Returns the release date and time of a Spotify album by its ID.\\tweet-day-extractor: Returns the calendar day a tweet was posted.\\temp-at-location: Returns current temperature for a specified address.\\historical-event-date: Returns a significant historical date for the specified location.\\mul: multiplies two values together\\spotify-album-finder: Returns Spotify album ID for a given artist or band name.\\add: sums two values together\\extract-year-from-datetime: Extracts the year from a given datetime input.\\movie-recommender: Suggests a similar movie based on the input title.\\airline-founding-date: Returns the founding date of the specified airline.\\\\Calculate the product of the release year of a Kaiser Chiefs album and the sum of that year with itself.\\\\\\\hl{<request><spotify-album-finder>`Kaiser Chiefs'<call>}27314933599224<response>\\\hl{<request><spotify-album-release-date>27314933599224<call>}`04:08 19/3/1720'\\<response>\\\hl{<request><extract-year-from-datetime>`04:08 19/3/1720'<call>}1458<response>\\\hl{<request><add>1458,1458<call>}2916<response>\\\hl{<request><mul>2916,1458<call>}4251528<response>\\\hl{Result=4251528<submit>}
    }%
}

\hspace{0mm}
\\

\hspace*{-3.85mm}\fbox{%
    \parbox{74mm}{\tiny
        The tools at your disposal:\\starbucks-item-suggester: Suggests a complementary Starbucks item based on the input item name.\\email-greeting-generator: Generates a personalized email template based on the recipient's name.\\starbucks-calorie-match: Suggests a Starbucks item based on the calorie range represented by the float.\\seasonal-recipe-suggester: Suggests a recipe based on seasonal ingredients available in the given month.\\rating-to-color: Maps Netflix ratings to a corresponding color for visual representation.\\starbucks-points-calculator: Returns reward points for a given Starbucks product ID.\\daylight-duration-estimator: Estimates average daylight hours for a given month.\\recipe-to-cuisine: Identifies the cuisine category based on the given recipe name.\\spotify-genre-finder: Returns the genre of a song using its Spotify ID.\\age-to-genre-mapper: Suggests music genre based on age demographic preferences.\\peak-tourism-month: Returns the peak tourism month for a given location.\\hashtag-peak-time: Returns peak engagement time for a given Twitter hashtag.\\age-calculator: Calculates age in years from a given historical datetime to the current date.\\cuisine-to-location: Suggests a city known for the specified cuisine.\\\\Suggest a complementary Starbucks item based on a seasonal recipe for October, considering the peak tourism month and average daylight hours of a city known for the recipe's cuisine.\\\\\\\hl{<request><seasonal-recipe-suggester>`October'<call>}`Tom Yum Soup'<response>\\\hl{<request><recipe-to-cuisine>`Tom Yum Soup'<call>}`Flemish'<response>\\\hl{<request><cuisine-to-location>`Flemish'<call>}`Jakarta'<response>\\\hl{<request><peak-tourism-month>`Jakarta'<call>}`March'<response>\\\hl{<request><daylight-duration-estimator>`March'<call>}2.2<response>\\\hl{<request><starbucks-calorie-match>2.2<call>}`Hazelnut Hot Chocolate'<response>\\\hl{<request><starbucks-item-suggester>`Hazelnut Hot Chocolate'<call>}`Peach Green Tea Lemonade'<response>\\\hl{Result=`Peach Green Tea Lemonade'<submit>}
    }%
}

\hspace{0mm}
\\

\hspace*{-3.85mm}\fbox{%
    \parbox{74mm}{\tiny
        The tools at your disposal:\\recipe-swapper: Suggests an alternative recipe based on the input recipe name.\\add: sums two values together\\ingredient-to-amazon-id: Maps ingredient names to corresponding Amazon product IDs for easy shopping.\\flight-price-checker: Retrieves the current ticket price for a specified flight ID.\\amazon-review-fetcher: Retrieves average review rating for a specified Amazon item by its ID.\\sub: subtracts the second argument from the first\\\\Calculate the negative of the average review rating for the Amazon item with ID 712523906030543.\\\\\\\hl{<request><amazon-review-fetcher>712523906030543<call>}3.1<response>\\\hl{<request><add>3.1,3.1<call>}6.2<response>\\\hl{<request><sub>3.1,6.2<call>}-3.1<response>\\\hl{Result=-3.1<submit>}
    }%
}

\section{Information Leakage}
\label{app_info_leakage}

To investigate the potential for tool names and/or descriptions leaking into instructions during the generation process described in Section \ref{sec_randomworld_sub_env_sub_instructions}, we manually analyzed 200 randomly-sampled task/instruction pairs from the RandomWorld train split of Section \ref{sec_experiments}. In particular, we flagged any substrings of the generated task instructions that were either similar or identical to the names or descriptions of tools in the trajectory skeleton provided to the instruction-generation LLM. 

Of the 200 instructions, we found potential information leakage in 133 instances (66.5\%). Examples are given below: suspicious instruction substrings and the corresponding substrings in the tool names/descriptions are \hl{highlighted}. \\

\hspace*{-3.85mm}\fbox{%
    \parbox{74mm}{\tiny
        \textbf{Tool Descriptions:} \\
        daily-trend-hashtag: Returns \hl{trending Twitter hashtag for the specified day of the week.} \\
        int-to-weekday: Converts an integer to a corresponding day of the week, modulo 7. \\
        month-to-number: Converts a month's name to its corresponding calendar number. \\
        hashtag-to-post-id: Fetches a recent post ID using the specified hashtag. \\\\
        \textbf{Instruction:} \\
        Find a recent Twitter post ID using \hl{the trending hashtag for the day of the week} corresponding to the month of `February'.
    }%
}

\hspace{0mm}
\\

\hspace*{-3.85mm}\fbox{%
    \parbox{74mm}{\tiny
        \textbf{Tool Descriptions:} \\
        tweet-activity-time: Returns the most active time for a given Twitter username. \\
        time-to-playlist: Returns a playlist ID based on the time of day for mood setting. \\
        hashtag-influencer-finder: Finds \hl{top influencer associated with a given hashtag.} \\
        trend-tag-locator: Returns trending Twitter hashtag for a specified location. \\
        genre-to-netflix-id: Returns a Netflix movie ID based on the specified genre. \\
        netflix-duration-lookup: Returns the duration in hours for a given Netflix movie ID. \\
        playlist-genre-identifier: Identifies the dominant genre of a Spotify playlist using its ID. \\
        add: sums two values together \\\\
        \textbf{Instruction:} \\
        Find the duration of a Netflix movie, based on the genre of a Spotify playlist that matches the most active time of a \hl{top influencer associated with a trending hashtag} in `Louisville', and then double that duration.
    }%
}

\hspace{0mm}
\\

\hspace*{-3.85mm}\fbox{%
    \parbox{74mm}{\tiny
        \textbf{Tool Descriptions:} \\
        future\_time\_calculator: Adds hour duration to current time, returning future date and time. \\
        datetime\_to\_month: Extracts and returns the month name from a given datetime input. \\
        driver\_shift\_duration: Returns the \hl{duration of the last completed shift for a given Uber driver.} \\
        div: divides the first argument by the second \\
        mul: multiplies two values together

        \textbf{Instruction:} \\
        Determine the month name of the current date and time after adding the \hl{duration of the last completed shift for Uber driver} with ID 963846425985.
    }%
}

\hspace{0mm}

\hspace{0mm}

Despite this admittedly high rate of information leakage, we again argue that the results of our RandomWorld-trained models in Table \ref{table_results} indicate that any information leakage that may have occurred did not substantially detract from the models’ performance. 

Furthermore, none of the examples that we manually inspected exhibited sufficient information leakage to entirely give away the correct tool call sequence. Note that the RandomWorld-trained models in Table \ref{table_results} have substantially higher performance on the withheld RandomWorld test set than those not trained on RandomWorld: Llama-RW-GRPO (0.721), Llama-RW-SFT (0.855), Qwen-RW-GRPO (0.536), and Qwen-RW-SFT (0.714) all outperform Llama-3.1-8B-Instruct (0.483), ToolACE-8B (0.170), Qwen2.5-7B-Instruct (0.109), Hammer2.0-7B (0.011), and GPT-3.5 (0.348) by a wide margin. We argue that the poor performance of the models not trained on RandomWorld indicates that any information leakage in the generated RandomWorld instructions did not render these tasks trivial. 

\section{GRPO Training Details}
\label{app_grpo}

As discussed in Section \ref{sec_randomworld}, we implement the policy-environment interaction via TRL text environments during GRPO training in Section \ref{sec_experiments}. The prompt/question $q$ consists of two few-shot examples, the tool inventory (and distractor tools), and the instruction. 

The policy-generated text and the tool outputs together (all orange and blue highlighted text in Figure \ref{fig_fig1}) are taken as the completion to $q$. However, we employ the TRL text environment's response masking feature, so that the GRPO loss is only calculated with respect to the policy-generated text (all orange text in Figure \ref{fig_fig1}).

As in the RandomWorld test set described in Section \ref{sec_experiments_sub_eval}, the policy model was limited to a maximum of 15 turns (tool calls or attempted tool calls) and 128 new tokens per turn before the trajectory is aborted, returning a reward of 0.0.

We trained all models (including SFT) on two H100 GPUs, with generation parallelization during GRPO training for speedup.

\end{document}